\pdfoutput=1
\documentclass[11pt]{article}
\usepackage[final]{acl}
\usepackage{times}
\usepackage{latexsym}
\usepackage[T1]{fontenc}
\usepackage[utf8]{inputenc}
\usepackage{microtype}
\usepackage{inconsolata}
\usepackage{graphicx}
\usepackage{booktabs}
\usepackage{amsmath}
\usepackage{amssymb}
\usepackage{xspace}
\usepackage{tabularx}
\usepackage{enumitem}
\usepackage{fvextra}
\usepackage{algorithm}
\usepackage{algpseudocode}
\usepackage[table]{xcolor}
\definecolor{posgreen}{RGB}{226, 246, 226}
\definecolor{negred}{RGB}{252, 226, 226}
\definecolor{lightblue}{RGB}{226, 240, 255}

\DefineVerbatimEnvironment{Prompt}{Verbatim}{
  fontsize=\scriptsize,
  breaklines=true,
  breakanywhere=true
}

\newcommand{\method}{DCCD\xspace}
\newcommand{\methodlong}{Dual-Confidence Contrastive Decoding\xspace}

\title{Dual-Confidence Contrastive Decoding for Retrieval-Augmented Generation}

\author{
  \textbf{Raymond Li$^{\spadesuit\diamondsuit}$ \quad
  Md Tawkat Islam Khondaker$^{\spadesuit\diamondsuit}$ \quad
  Amirhossein Abaskohi$^{\diamondsuit}$} \\[2pt]
  \textbf{Gabriel Murray$^{\diamondsuit}$ \quad
  Giuseppe Carenini$^{\diamondsuit}$ \quad
  Issam H. Laradji$^{\spadesuit\diamondsuit}$} \\[4pt]
  $^{\spadesuit}$ServiceNow Research \quad
  $^{\diamondsuit}$University of British Columbia
}

\begin{document}
\maketitle

\begin{abstract}
Retrieval-augmented generation (RAG) often requires models to answer questions from multiple retrieved documents, where only some sources are relevant and the retrieved bundle may contain stale, noisy, or conflicting evidence. Existing contrastive decoding methods primarily focus on resolving conflicts between the model's internal memory and the retrieved context. In contrast, we study the complementary problem of intra-context conflict in multi-document RAG. To evaluate this setting, we introduce \textbf{DRQA}, a factual-conflict question answering benchmark derived from enterprise deep-research scenarios, where answers are grounded in synthetic enterprise-specific facts that are designed not to be recoverable from the model's internal memory.
We further propose \textbf{Dual-Confidence Contrastive Decoding} (\textbf{DCCD}), a training-free decoding method that combines document-level confidence, which estimates whether a document appears sufficient for answering the question, with token-level confidence, which estimates whether that document supports a confident next-token prediction. Across DRQA and standard multi-document QA benchmarks, DCCD achieves the best average performance among full-context and contrastive decoding baselines, with the largest gains on DRQA. These results highlight the importance of source-aware, confidence-gated decoding when retrieved evidence is internally conflicting.\footnote{\url{https://github.com/ServiceNow/dccd-rag}}
\end{abstract}

\section{Introduction}
\label{sec:introduction}

Retrieval-augmented generation (RAG) has become an essential paradigm for grounding large language models (LLMs) in external knowledge \citep{gao2023retrieval}, enabling them to answer domain-specific and knowledge-intensive queries beyond what is stored in their parameters. Initially developed for open-domain question answering (QA) \citep{karpukhin-etal-2020-dense, lewis2020retrieval}, RAG has become an integral component of workflows that require resolving information from multiple sources, from open-domain deep-research agents that synthesize evidence across the web \citep{du2026deepresearch, java2026characterizing} to enterprise systems that retrieve from heterogeneous data such as reports, emails, and chats \citep{choubey-etal-2025-benchmarking, abaskohi2026drbench}.

Despite these advancements, a central challenge for RAG systems is
the resolution of conflicts among the knowledge sources \citep{xu-etal-2024-knowledge-conflicts}. At inference time, the LLM can be overly reliant on its internal memory acquired during pre-training, even when the retrieved context provides newer or more task-specific evidence \citep{chen-etal-2022-rich, xie2024adaptive}. One popular training-free approach to context-memory conflict is contrastive decoding (CD) \citep{li-etal-2023-contrastive}, which modulates generation by contrasting model predictions under different conditioning contexts \citep{shi-etal-2024-trusting, wang-etal-2025-adacad, khandelwal-etal-2025-cocoa}. Rather than fine-tuning the LLMs, contrastive decoding adjusts token probabilities based on the difference between output distributions with and without the context. However, this formulation treats the retrieved evidence as a single context and does not resolve conflicts within the retrieved context itself.

In multi-document QA, the model needs to decide the relevance of each document to the query and which evidence to use \citep{wan-etal-2024-evidence, jin-etal-2024-tug}. The retrieved evidence may contain answer-supporting evidence alongside irrelevant distractors, stale information, and plausible but incorrect claims.
In this setting, the decoding problem is not only about choosing the next token, but also about choosing the evidence source that should inform that token. We use \emph{confidence} as the complement of \emph{uncertainty} and distinguish two levels: \emph{document-level confidence}, which estimates whether a retrieved document appears sufficient for answering the question, and \emph{token-level confidence}, which estimates whether a document-conditioned stream yields a sharp local next-token distribution.
Token-level confidence has been used to capture how confidently the model predicts under a given context \citep{malinin2021uncertainty, jiang-etal-2023-active, fadeeva-etal-2024-fact, nguyen2026probabilities}, but it cannot determine whether the context itself is sufficient to answer the question. Document-level confidence addresses the complementary question of which retrieved source contains answer-bearing evidence \citep{kadavath2022language, singh-etal-2023-tree, wan-etal-2024-evidence, joren2025sufficient}, but it does not ensure that the model can generate from that source with confidence. Effective multi-document RAG therefore requires both source selection and generation confidence.

Most existing RAG and conflict-resolution benchmarks do not directly isolate this joint capability. For example, retrieval-augmented QA benchmarks \citep{joshi-etal-2017-triviaqa, kwiatkowski-etal-2019-natural, mallen-etal-2023-trust, zhang-etal-2024-retrievalqa} mainly test whether models can ground answers in retrieved context despite noisy retrievals, while conflict-resolution benchmarks \citep{leeambigdocs, wangretrieval} introduce contradictory evidence through altered public facts or ambiguous entity references, such that resolving the conflict may depend on the model's internal world knowledge rather than on confidence over the retrieved evidence itself. A model that already knows the true fact can answer correctly without arbitrating among the retrieved documents, so existing benchmarks cannot distinguish evidence-based arbitration from reliance on internal memory \citep{mallen-etal-2023-trust, xie2024adaptive}. This challenge is especially pronounced in enterprise settings, where relevant facts are often private or organization-specific, and retrieved evidence can be heterogeneous, time-sensitive, and internally conflicting \citep{choubey-etal-2025-benchmarking, abaskohi2026drbench}.

Motivated by these gaps, we introduce an enterprise benchmark for multi-document question answering and propose a novel decoding method. To address the \textit{evaluation gap}, we propose \textbf{DRQA}, a factual-conflict QA benchmark derived from the recent DRBench \citep{abaskohi2026drbench} consisting of synthetic enterprise-specific facts grounded in company profiles, user personas, and deep-research tasks, where the answer cannot be inferred from the model's internal memory. Instead, each answer must be recovered from a retrieved bundle containing correct, stale temporal, misinformation, and noise documents, requiring models to identify the evidence source that supports the answer. To address the \textit{decoding gap}, we propose \methodlong~(\textbf{\method}), a training-free contrastive decoding method that uses dual-confidence signals: document-level confidence estimates whether a document appears sufficient for answering the question, while token-level confidence estimates whether the model can produce a sharp next-token prediction from that document. \method uses these signals to select positive and negative document-conditioned streams and scales the contrastive term by their confidence margin, amplifying high-confidence evidence while suppressing conflicting or unsupported alternatives. Across multiple RAG benchmarks, \method achieves the best average performance, with the largest gains on DRQA.

Our contributions are threefold.
(1) We introduce \textbf{DRQA}, a factual-conflict QA benchmark derived from enterprise deep-research scenarios, where each query is paired with a retrieved bundle containing one correct document and distractors that are misinformation, stale temporal, or noise.
(2) We propose \methodlong~(\textbf{\method}), a training-free contrastive decoding method that combines document-level answerability confidence with token-level generation confidence to perform confidence-gated document-level contrast.
(3) We show that \method achieves the best result in 28 of 30 settings across DRQA and standard multi-document QA benchmarks, with especially large gains on DRQA, highlighting the importance of source-aware contrast when retrieved evidence is internally conflicting.

\section{Related Work}
\label{sec:related-work}

\subsection{Retrieval-Augmented Question Answering}
Despite their capabilities, LLMs remain prone to hallucination and are limited by the knowledge acquired during training, making it necessary to ground generation in verifiable external evidence. Retrieval-augmented generation (RAG) dynamically incorporates information through dense passage retrieval \citep{karpukhin-etal-2020-dense}, and has been shown to improve open-domain question answering when the model's internal knowledge is insufficient \citep{lewis2020retrieval}.
Consequently, models are typically evaluated on open-domain QA benchmarks on their ability to answer factual questions using retrieved passages from public corpora. For example, \citet{izacard-grave-2021-leveraging} use real Google queries and Wikipedia answers to evaluate the ability of the model to ground responses in retrieved evidence, PopQA~\citep{mallen-etal-2023-trust} further probes the boundary between internal and retrieved knowledge by focusing on long-tail facts for which retrieval is often necessary, while RetrievalQA~\citep{zhang-etal-2024-retrievalqa} evaluates adaptive RAG behavior under retrieved contexts of varying usefulness.

Since knowledge conflict can also arise within the retrieved context itself \citep{xu-etal-2024-knowledge-conflicts}, conflict-oriented benchmarks have been proposed to evaluate the models' ability to arbitrate among conflicting evidence. However, existing benchmarks are constructed by altering public facts or referencing ambiguous entities \citep{leeambigdocs, wangretrieval}.
In these settings, models may still be able to resolve the conflict using internal world knowledge (e.g., recognizing the true public fact or the intended entity) rather than by reasoning over the retrieved text itself.
While a multiple-choice benchmark built on conflicts over fictional entities avoids this reliance on world knowledge \citep{schuster2026whose}, DRQA evaluates free-form generation from multi-document enterprise evidence.
In parallel, deep-research benchmarks evaluate broader systems that gather and synthesize evidence across many sources, including open-domain web research and enterprise settings \citep{du2026deepresearch, java2026characterizing, abaskohi2026drbench}.
Our proposed \textbf{DRQA} dataset fills this gap by
combining conflict-oriented factual QA with enterprise deep-research scenarios, where the answer is unlikely to be available from internal memory and must instead be recovered by identifying which retrieved document supports the answer.

\subsection{Contrastive Decoding}
Contrastive decoding (CD) \citep{li-etal-2023-contrastive} was originally proposed as a training-free strategy that selects tokens with high expert-model probability and low amateur-model probability, using their probability discrepancy to favor fluent and informative continuations. Context-Aware Decoding (CAD) \citep{shi-etal-2024-trusting} adapts this idea to improve context faithfulness in RAG by contrasting the LLM with and without retrieved context, so that generation follows retrieved evidence rather than prior knowledge. Further work makes the contrastive weight adaptive \citep{zhao-etal-2024-enhancing, kim-etal-2024-adaptive}. For example, \textsc{AdaCAD} \citep{wang-etal-2025-adacad} adjusts the contrastive weight at each decoding step using the Jensen-Shannon divergence (JSD) between contextual and zero-shot distributions, while \textsc{CoCoA} \citep{khandelwal-etal-2025-cocoa} introduces token-level confidence-aware adaptation to address conflict between contextual and internal knowledge.
Most similar to our work is Dynamic Contrastive Decoding (DVD) proposed by \citet{jin-etal-2024-dvd}, which targets multi-document QA by contrasting document-conditioned predictions and dynamically increasing the influence of documents estimated to be useful for the current generation step.
However, DVD does not explicitly distinguish document-level support from token-level generation confidence.

\subsection{Confidence Estimation}
Current work on confidence estimation has primarily focused on confidence in the model's output distribution \citep{geng-etal-2024-survey}. Entropy-based methods estimate predictive uncertainty for autoregressive generation \citep{malinin2021uncertainty, jiang-etal-2023-active}, and token-level uncertainty has been used for hallucination detection and fact checking \citep{fadeeva-etal-2024-fact}. 
Recent probability-only methods motivate lightweight top-\(k\) token statistics for uncertainty estimation \citep{nguyen2026probabilities}, where our token-level confidence follows the Dirichlet uncertainty framework of \citet{malinin-gales-2018-predictive} by mapping the top-\(k\) logits of each stream to Dirichlet concentration parameters and computing the expected entropy.
In parallel, document-level confidence addresses the complementary problem of whether an evidence source is sufficient to answer the question.
Prior work has studied the models' ability to self-evaluate whether they know an answer \citep{kadavath2022language}, identify convincing evidence under conflicting sources \citep{wan-etal-2024-evidence}, and estimate the sufficiency of sources in retrieved evidence \citep{joren2025sufficient, hwang-etal-2025-retrieval}. The closest work to our document-level confidence is context-sufficiency estimation \citep{joren2025sufficient}, which asks whether the provided context contains enough information to answer a question. In this work, our proposed \method combines document-level and token-level confidence estimates inside the contrastive decoding process.

\section{DRQA: Deep Research Question Answering Benchmark}
\label{sec:drqa}

We derive \textbf{DRQA} from DRBench~\citep{abaskohi2026drbench}, a 100-task enterprise deep-research benchmark spanning three industries---retail, healthcare, and electric vehicles---and ten domains. DRBench contains internal and external supporting facts embedded in generated enterprise documents verified by human annotators. We use the 568 synthetically generated internal facts, which are designed not to appear on the public web and therefore are not expected to be recoverable from the model's internal memory. Additional construction details and prompts are provided in \autoref{app:drqa-details}.

\subsection{Dataset Construction}

Each DRBench task contains internal insights generated from the organization's industry, priorities, customer segments, and business goals. Since we focus on retrieval-augmented question answering rather than full deep-research report generation, we use the question from the DRBench internal insights. For example, the insight ``Lee's Market reduced food waste by 8\% in Q2 2024, saving \$1.2M'' comes with the question ``What is Lee's Market's current food waste reduction rate as of Q2 2024?''

For each QA pair, we construct an evidence set with four document types: \emph{correct}, \emph{misinformation}, \emph{temporal}, and \emph{noise}. Correct documents contain the ground-truth answer, misinformation documents contain a plausible but incorrect answer, temporal documents contain an answer that was valid at an earlier document date, and noise documents are topically plausible but do not answer the question. Each document is assigned a document date in ISO 8601 format and a source format, which is email, chat, or Markdown (Appendix~\ref{app:pptx-reconstruction}).

We reuse the DRBench insight and distractor files, which are generated with a three-step needle-in-a-haystack procedure~\citep{abaskohi2026drbench} (Appendix~\ref{app:drbench-background}), as correct and noise documents, and serialize all documents as Markdown or JSONL for model input.

\paragraph{Misinformation}
To construct \emph{misinformation} documents, we prompt the LLM to generate a plausible perturbation of the ground-truth insight and then render it with a format-specific generator (Appendix~\ref{app:gen-prompts}). To make these documents realistic but not explicitly labeled, we add subtle format-specific uncertainty markers, such as conversational hedges in chat (``pretty sure'', ``last I heard'') or tentative phrasing in reports (``preliminary findings'', ``appears to''). These markers reflect the hedging language that occurs naturally in enterprise communication, but do not identify the document as incorrect. Thus, models must rely on weak cues, source metadata, and context rather than explicit correctness labels.

\paragraph{Temporal}
Temporal documents are generated to read as \emph{authoritative at the time of writing}, rather than as retrospective historical summaries. We first sample the \texttt{document\_date} uniformly between 6 months and 3 years before the generation date (correct documents are dated 1 week to 3 months before it), then prompt the LLM to generate a value that would have been accurate and current as of that date. A second call produces the full document in present tense using the same format-specific generators.

We ensure that temporal documents contain no explicit ``archived'' or ``outdated'' labels. Their staleness shows only in the \texttt{document\_date} and in the earlier period that the stated value refers to. The direction of change is unconstrained, so a stale value cannot be spotted by judging whether a trend is intuitive. Document dates and source formats are kept as metadata and are not part of the model input, although generated documents may mention dates in their body (Appendix~\ref{app:bundle-stats}).

\subsection{Quality Verification}
\label{sec:drqa-verify}

Complementing DRBench's per-file human spot-check protocol, we apply an automated LLM-as-a-judge quality filter to all generated conflict documents. Each document type is checked against a fixed rubric that returns a structured JSON binary verdict. For misinformation documents, we verify that the claimed answer is factually and specifically different from the ground truth while remaining plausible to an uninformed reader. For temporal documents, we verify that the historical answer is distinct from the current ground truth and that the document contains no explicit staleness labels. Failing documents are regenerated and re-checked, and we manually inspect the remaining failures.

Unlike standard open-domain QA benchmarks, where retrieval is usually expected to help, DRQA creates a setting in which retrieval is necessary but not sufficient. The answers are synthetic private facts, and each retrieved bundle contains both the correct answer and plausible contradictions. The evaluation question is therefore not only whether the model benefits from retrieval, but which value it commits to when confronted with conflicting and noisy evidence. DRQA directly evaluates retrieval-grounded conflict resolution, a capability that remains underexplored in prior work.

\subsection[Scope and Conflict Types]{Scope and Conflict Types}
\label{sec:drqa-scope}
Each DRQA example has exactly one current gold document. In enterprise deep research, an insight usually comes from a single authoritative document, and a report combines many such insights \citep{choubey-etal-2025-benchmarking, abaskohi2026drbench}. DRQA therefore leaves out questions that require combining several documents. Its distractors cover factual, temporal, and irrelevance conflicts, but not authority, scope, partial-context, or interpretation conflicts (Appendix~\ref{app:conflict-taxonomy}).

\section{Methodology}
\label{sec:method}
\begin{figure*}[t]
    \centering
    \includegraphics[width=\textwidth]{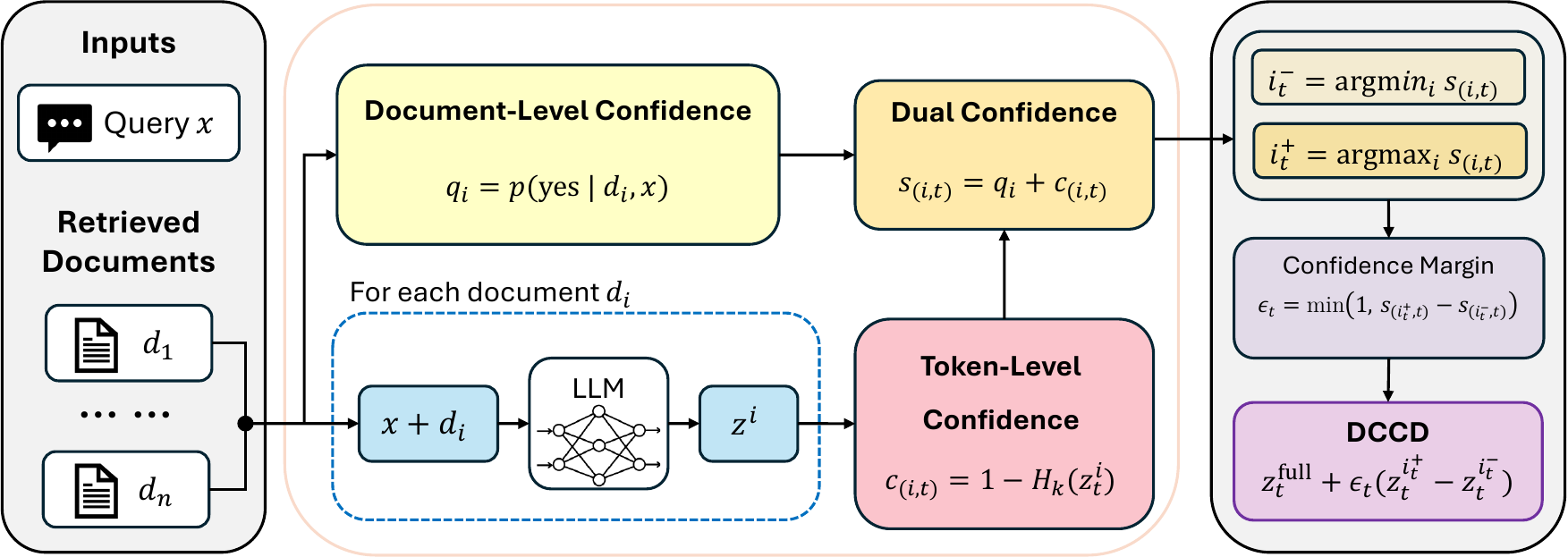}%
    \caption{Overview of \methodlong (\method). Given a query \(x\) and retrieved documents \(\{d_1,\ldots,d_n\}\), $z^i$ denotes the next-token logits conditioned on document $d_i$. \method combines document-level (\(q_i\)) and token-level (\(c_{(i,t)}\)) confidence scores to select positive and negative document-conditioned streams at each decoding step. The margin between the selected scores, clipped at 1, controls the strength of the contrastive term \(\epsilon_t\). }
    \label{fig:dccd}
\end{figure*}

In this section, we describe our proposed \methodlong~(\textbf{\method}) with an overview diagram illustrated in \autoref{fig:dccd}. Additional derivations and implementation details of \method are presented in \autoref{app:dccd}.

\subsection{Problem Setup}
\label{sec:problem-setup}

Let \(x\) denote the question and let
$\mathcal{D}=\{d_1,\ldots,d_n\}$
be the list of documents returned by the retriever. The goal is to generate an accurate and precise answer to \(x\) using \(\mathcal{D}\). At each decoding step \(t\), the model produces next-token logits \(\mathbf{z}_t\) conditioned on the current prefix \(y_{<t}\), and the decoding strategy selects the next token \(y_t\).

\subsection{Document-Level Confidence}
\label{sec:document-confidence}

In multi-document RAG, the retrieved set $\mathcal{D}$ can contain documents with different levels of relevance and evidential value. While retrieval similarity scores provide a useful filter, they are optimized for query-document relevance rather than for whether a document contains enough information to answer the question. Motivated by prior work studying whether the model knows the answer \citep{kadavath2022language} and the sufficiency of a retrieved document for answering the query $x$ \citep{wan-etal-2024-evidence, joren2025sufficient}, we first estimate a document-level confidence score for each question-document pair $(x, d_i)$.

Specifically, we construct a support-probe prompt $\pi_{\mathrm{sup}}(x,d_i)$ that asks the model whether $d_i$ contains enough information to answer $x$. The probe is formulated as a binary yes/no question. Let $\mathcal{Y}_{+}$ be the set of yes verbalizer tokens and $\mathcal{Y}_{-}$ be the set of no verbalizer tokens, including common case and leading-space variants. We use the model's next-token distribution after the support probe, and define the document-level confidence score $q_i$ as in \autoref{eq:q-answerability}.

\begin{equation}
q_i =
\frac{
\sum_{v\in\mathcal{Y}_{+}}
p_\theta(v\mid \pi_{\mathrm{sup}}(x,d_i))
}{
\sum_{v\in\mathcal{Y}_{+}\cup\mathcal{Y}_{-}}
p_\theta(v\mid \pi_{\mathrm{sup}}(x,d_i))
}.
\label{eq:q-answerability}
\end{equation}

Document-level confidence $q_i$ is computed once before decoding and estimates whether a retrieved document $d_i$ should be considered a candidate evidence source. Importantly, the support probe requires only a single forward pass per document, and adds a one-time preprocessing cost that is amortized over the answer-generation steps.

\subsection{Token-Level Confidence}
\label{sec:token-confidence}

Document-level confidence estimates whether a retrieved document is sufficient for answering the question, but it does not indicate whether the model can use that document to make a confident prediction at a particular decoding step. We therefore compute a token-level confidence score for each document-conditioned generation stream.

At decoding step \(t\), let \(\mathbf{z}^i_t \in \mathbb{R}^{V}\) denote the next-token logits produced by the model conditioned on the question \(x\), document \(d_i\), and the current prefix \(y_{<t}\). Following the Dirichlet uncertainty framework of \citet{malinin-gales-2018-predictive}, we approximate predictive uncertainty by treating the top-\(k\) logits of each document-conditioned stream as evidence for a Dirichlet distribution over next-token categorical distributions. This gives a lightweight, training-free confidence estimate based only on the most relevant next-token alternatives \citep{nguyen2026probabilities}.

Let \(\mathcal{T}_{(i,t)}\) be the set of top-\(k\) token indices under \(\mathbf{z}^i_t\). For each \(v\in\mathcal{T}_{(i,t)}\), we define the Dirichlet concentration parameter:
\begin{equation}
\alpha_{(i,t,v)}
=
\operatorname{softplus}\!\left(\mathbf{z}^i_t[v]\right)+1.
\label{eq:token-alpha}
\end{equation}
Let:
\begin{equation}
S_{(i,t)}
=
\sum_{v\in\mathcal{T}_{(i,t)}}
\alpha_{(i,t,v)},~~~
\bar p_{(i,t,v)}
=
\frac{\alpha_{(i,t,v)}}{S_{(i,t)}}
\label{eq:token-dirichlet-mean}
\end{equation}
We measure top-\(k\) uncertainty as the expected entropy of the categorical distribution induced by this Dirichlet posterior:
\begin{equation}
H_k(\mathbf{z}^i_t)
=
\sum_{v\in\mathcal{T}_{i,t}}
\bar p_{i,t,v}
\bigl[
\psi(S_{i,t}+1)
-
\psi(\alpha_{i,t,v}+1)
\bigr]
\label{eq:token-raw-entropy}
\end{equation}%
where \(\psi(u)=\frac{d}{du}\log\Gamma(u)\) is the digamma function, so that \(0\le H_k\le\log k\).

We define token-level confidence as
\begin{equation}
c_{(i,t)}
=
1-H_k(\mathbf{z}^i_t).
\label{eq:token-confidence}
\end{equation}

Thus, \(c_{(i,t)}\) is high when document \(d_i\) yields a sharp next-token prediction under the current prefix, and low when its evidence is spread across many plausible next tokens. Unlike \(q_i\), which is computed once before decoding, \(c_{(i,t)}\) is recomputed at every decoding step and captures whether a candidate evidence source supports a clear local continuation.
\subsection{Dual-Confidence Contrastive Decoding}
\label{sec:dccd}

At each decoding step, \method combines document-level and token-level confidence to choose the document-conditioned streams that define the contrastive direction. For each document $d_i$, we define the dual-confidence score as the sum of the two terms:
\begin{equation}
s_{(i,t)}
=
q_i + c_{(i,t)}.
\label{eq:dual-confidence}
\end{equation}
This unweighted sum provides a simple hyper\-parameter-free aggregation of source sufficiency and local generation confidence, where weighted extensions are left as future work.

From \autoref{eq:dual-confidence}, we select the positive and negative document-conditioned streams based on the highest and lowest dual-confidence scores at each decoding step:
\begin{equation}
i_t^+ = \arg\max\nolimits_{i} s_{(i,t)},
\quad
i_t^- = \arg\min\nolimits_{i} s_{(i,t)}.
\label{eq:document-selection}
\end{equation}
We use the margin between the positive and negative scores, clipped at 1, to control the strength of the contrastive term:
\begin{equation}
\epsilon_t
=
\min\bigl(1,\;
s_{(i_t^+,t)}
-
s_{(i_t^-,t)}\bigr).
\label{eq:epsilon}
\end{equation}%

Finally, let $\mathbf{z}^{\mathrm{full}}_t$ denote the next-token logits conditioned on the question with the full retrieved document set $\mathcal{D}$ and the current output prefix $y_{<t}$. Let $\mathbf{z}^{i}_t$ denote the next-token logits conditioned on the question, document $d_i$, and the same prefix. The decoding rule is then:
\begin{equation}
\mathbf{z}_t
=
\mathbf{z}^{\mathrm{full}}_t
+
\epsilon_t
\left(
\mathbf{z}^{i_t^+}_t
-
\mathbf{z}^{i_t^-}_t
\right).
\label{eq:dccd-rule}
\end{equation}
This rule anchors generation in the full retrieved context by adding a document-level contrast that amplifies evidence from the most confident source and suppresses evidence from the least confident source at each decoding step.
Since $s_{(i,t)}$ is recomputed at every step, the selected streams can change during generation, while the full-context anchor keeps every document available. Algorithm~\ref{alg:dccd} summarizes the procedure.

\begin{algorithm*}[t]
\caption{\method decoding for one query.}
\label{alg:dccd}
\begin{algorithmic}[1]
\Require question $x$, documents $d_{1:n}$, top-$k$ size $k$
\State $q_i \gets$ support probe on $d_i$, for all $i$ \Comment{Eq.~\eqref{eq:q-answerability}}
\For{$t=1,2,\ldots$}
  \State $\mathbf{z}^{\mathrm{full}}_t,\mathbf{z}^{1:n}_t \gets$ next-token logits \Comment{batched}
  \State $s_{(i,t)} \gets q_i + 1 - H_k(\mathbf{z}^i_t)$, for all $i$ \Comment{Eq.~\eqref{eq:dual-confidence}}
  \State $i^+_t \gets \arg\max_i s_{(i,t)}$,\; $i^-_t \gets \arg\min_i s_{(i,t)}$
  \State $\epsilon_t \gets \min\bigl(1,\, s_{(i^+_t,t)} - s_{(i^-_t,t)}\bigr)$ \Comment{Eq.~\eqref{eq:epsilon}}
  \State $y_t \gets \arg\max_v \bigl[\mathbf{z}^{\mathrm{full}}_t + \epsilon_t(\mathbf{z}^{i^+_t}_t - \mathbf{z}^{i^-_t}_t)\bigr]_v$
  \State \textbf{stop} at end-of-sequence or a complete first sentence
\EndFor
\end{algorithmic}
\end{algorithm*}

\section{Experiments}
\label{sec:experiments}

\begin{table*}[t]
\centering
\resizebox{\linewidth}{!}{%
\begin{tabular}{l|cc|cc|cc|cc|cc}
\toprule
& \multicolumn{2}{c|}{\textbf{DRQA}} & \multicolumn{2}{c|}{\textbf{NQ}} & \multicolumn{2}{c|}{\textbf{TriviaQA}} & \multicolumn{2}{c|}{\textbf{RetrievalQA}} & \multicolumn{2}{c}{\textbf{PopQA}} \\
\cmidrule(lr){2-3}\cmidrule(lr){4-5}\cmidrule(lr){6-7}\cmidrule(lr){8-9}\cmidrule(lr){10-11}
Method & @5 & @10 & @5 & @10 & @5 & @10 & @5 & @10 & @5 & @10 \\
\midrule
\multicolumn{11}{c}{\texttt{Qwen3.5-2B}} \\
\midrule
zero-shot & \multicolumn{2}{c|}{0.00} & \multicolumn{2}{c|}{15.62} & \multicolumn{2}{c|}{32.21} & \multicolumn{2}{c|}{47.83} & \multicolumn{2}{c}{18.44} \\
full   & 11.44 & 7.39 & 51.75 & 52.44 & 69.59 & 71.33 & 65.35 & 66.00 & 51.35 & 53.32 \\
CAD    & 10.92 & 7.39 & 51.30 & 52.22 & 69.75 & 71.56 & 65.24 & 66.14 & 51.30 & 52.61 \\
\textsc{AdaCAD} & 10.92 & 7.92 & 51.05 & 52.05 & 69.33 & 71.28 & 65.06 & 66.03 & 51.11 & 52.57 \\
DVD    & 12.68 & 6.87 & 50.25 & 50.30 & 69.57 & 71.13 & 64.56 & 65.21 & 50.41 & 52.31 \\
\textsc{CoCoA}  &  7.75 & 4.75 & 45.40 & 45.29 & 66.47 & 67.71 & 61.62 & 61.87 & 49.29 & 49.94 \\
\textbf{\method (ours)} & \cellcolor{lightblue}\textbf{16.73} & \cellcolor{lightblue}\textbf{12.15} & \cellcolor{lightblue}\textbf{53.10} & \cellcolor{lightblue}\textbf{53.24} & \cellcolor{lightblue}\textbf{70.04} & \cellcolor{lightblue}\textbf{72.01} & \cellcolor{lightblue}\textbf{66.25} & \cellcolor{lightblue}\textbf{66.61} & \cellcolor{lightblue}\textbf{51.75} & \cellcolor{lightblue}\textbf{53.34} \\
\midrule
\multicolumn{11}{c}{\texttt{Qwen3.5-9B}} \\
\midrule
zero-shot & \multicolumn{2}{c|}{0.00} & \multicolumn{2}{c|}{29.56} & \multicolumn{2}{c|}{57.76} & \multicolumn{2}{c|}{50.02} & \multicolumn{2}{c}{26.06} \\
full   & 17.78 & 10.56 & 54.99 & 57.23 & 72.47 & 74.30 & 65.67 & 66.10 & 50.93 & 52.10 \\
CAD    & 17.78 & 10.21 & 55.21 & 57.56 & 72.39 & 74.31 & 65.42 & 66.28 & 50.57 & 51.56 \\
\textsc{AdaCAD} & 16.73 & 10.74 & 55.04 & 57.70 & 72.46 & 74.35 & 65.21 & 66.03 & 50.50 & 51.49 \\
DVD    & 16.90 & 10.92 & 54.32 & 55.57 & 71.00 & 72.44 & 65.35 & 64.99 & 50.36 & 50.60 \\
\textsc{CoCoA}  & 14.61 & 10.74 & 50.83 & 51.61 & 71.35 & 73.50 & 64.34 & 64.81 & 51.16 & 52.21 \\
\textbf{\method (ours)} & \cellcolor{lightblue}\textbf{21.48} & \cellcolor{lightblue}\textbf{13.03} & \cellcolor{lightblue}\textbf{55.68} & \cellcolor{lightblue}\textbf{58.42} & \cellcolor{lightblue}\textbf{73.08} & \cellcolor{lightblue}\textbf{75.10} & \cellcolor{lightblue}\textbf{66.61} & \cellcolor{lightblue}\textbf{66.79} & \cellcolor{lightblue}\textbf{51.42} & \cellcolor{lightblue}\textbf{53.29} \\
\midrule
\multicolumn{11}{c}{\texttt{Phi-3-medium} (14B)} \\
\midrule
zero-shot & \multicolumn{2}{c|}{0.00} & \multicolumn{2}{c|}{39.11} & \multicolumn{2}{c|}{69.11} & \multicolumn{2}{c|}{49.08} & \multicolumn{2}{c}{37.22} \\
full   & 14.96 & 10.92 & 57.89 & 59.56 & 75.17 & 76.66 & 66.21 & 67.18 & 50.90 & 51.86 \\
CAD    & 15.49 & 10.56 & 58.06 & \cellcolor{lightblue}\textbf{59.78} & 75.06 & 76.60 & 66.10 & 67.00 & 50.74 & 51.68 \\
\textsc{AdaCAD} & 14.61 & 10.74 & 57.95 & 59.72 & 75.04 & 76.58 & 66.18 & 66.61 & 50.93 & 51.56 \\
DVD    & 12.85 &  7.75 & 57.56 & 58.81 & 75.06 & 76.10 & 65.46 & 66.54 & 50.18 & 50.39 \\
\textsc{CoCoA}  & 12.85 & 10.56 & 55.10 & 54.79 & \cellcolor{lightblue}\textbf{75.97} & 76.52 & 66.00 & 66.54 & 50.71 & 51.16 \\
\textbf{\method (ours)} & \cellcolor{lightblue}\textbf{20.25} & \cellcolor{lightblue}\textbf{12.32} & \cellcolor{lightblue}\textbf{58.31} & 59.56 & 75.47 & \cellcolor{lightblue}\textbf{77.06} & \cellcolor{lightblue}\textbf{67.86} & \cellcolor{lightblue}\textbf{67.94} & \cellcolor{lightblue}\textbf{51.39} & \cellcolor{lightblue}\textbf{52.14} \\
\bottomrule
\end{tabular}
}
\caption{Experiment results for \method and baseline decoding strategies with \texttt{Qwen3.5-2B}, \texttt{Qwen3.5-9B} and \texttt{Phi-3-medium} (14B). We report the \texttt{str-em} accuracy for NQ, TriviaQA, RetrievalQA, PopQA, and LLM-as-a-judge accuracy for DRQA for top-5 (@5) and top-10 (@10) documents provided by the respective datasets. The best result for each setting is \textbf{bolded}.}
\label{tab:results}
\end{table*}

\subsection{Settings}
\label{sec:settings}

\paragraph{Datasets}
In addition to DRQA, we evaluate on well-known multi-document question answering benchmarks, including Natural Questions (NQ)~\citep{kwiatkowski-etal-2019-natural} and TriviaQA~\citep{joshi-etal-2017-triviaqa}, using the preprocessing of \citet{izacard-grave-2021-leveraging}. We also evaluate on the more recent PopQA~\citep{mallen-etal-2023-trust} and RetrievalQA~\citep{zhang-etal-2024-retrievalqa}. For DRQA, since the ground-truth answer is expressed as free-form text, we use LLM-as-a-judge to evaluate semantic correctness against the ground-truth insight. For all other datasets, we use normalized exact string match (\texttt{str-em}) following prior work. 

\paragraph{Baselines}
We compare \method against standard full-context decoding (full) and contrastive decoding baselines. In particular, we include the standard CAD~\citep{shi-etal-2024-trusting}, DVD~\citep{jin-etal-2024-dvd}, \textsc{AdaCAD}~\citep{wang-etal-2025-adacad} and \textsc{CoCoA}~\citep{khandelwal-etal-2025-cocoa}. We also include the zero-shot baseline without any additional evidence for reference. Further details and hyperparameter settings of all decoding methods are described in \autoref{app:experiments}.

\paragraph{Models} Motivated by the recent trend of deploying Small Language Models (SLMs) for deep-research agents \citep{zheng-etal-2025-deepresearcher, wan2025pokeeresearch}, we evaluate all decoding strategies on instruction-tuned SLMs of small to medium sizes. Specifically, we use \texttt{Qwen3.5-2B}, \texttt{Qwen3.5-9B} \citep{qwen35blogcr}, and \texttt{Phi-3-medium} (14B) \citep{abdin2024phi3} provided by Hugging Face's \texttt{transformers} library \citep{wolf-etal-2020-transformers}. For all models, we use the same retrieval inputs and the default chat template used for instruction-training.

\subsection{Results}
As shown in \autoref{tab:results}, \method achieves the best result in 28 of the 30 settings. The gains hold across model sizes and datasets, with two exceptions for \texttt{Phi-3-medium} (NQ@10 and TriviaQA@5), and are especially pronounced on DRQA. They are smaller on the Wikipedia-style benchmarks, which by construction contain little intra-context conflict. We attribute these improvements to the dual-confidence design, where document-level confidence helps identify documents that appear sufficient for answering the question, while token-level confidence identifies document-conditioned streams that support sharp local predictions. These two signals address complementary failure modes. A document may appear answerable but still yield an uncertain next-token distribution, while a document may yield a sharp prediction despite supporting an incorrect or misleading answer.

On DRQA, zero-shot accuracy is \(0\%\) for all evaluated models, which is consistent with the benchmark design. Results on DRQA also show substantially lower accuracies than the Wikipedia-style datasets, as the retrieved evidence contains the correct document alongside noise, misinformation, and temporally stale evidence. We further observe an accuracy drop from @5 to @10 for every method, since each example contains only one current ground-truth document and @10 more than doubles the distractors (from four to nine). \method nevertheless remains the best method at @10 for all three models. In this conflict-heavy setting, \method achieves larger gains over baselines than on NQ, PopQA, RetrievalQA, and TriviaQA. This trend is consistent with the dual-confidence design, which favors document-conditioned streams that are both answerable and locally confident while contrasting against unsupported, stale, or misleading alternatives.

\subsection{Ablation Study}
\label{sec:ablations}

\begin{table}
\begin{center}
\setlength{\tabcolsep}{4pt}
\resizebox{\columnwidth}{!}{%
\begin{tabular}{l|c|c|c|c|c}
\toprule
Ablation & \textbf{DRQA} & \textbf{NQ} & \textbf{TQA} & \textbf{RetQA} & \textbf{PopQA} \\
\midrule
\multicolumn{6}{c}{@5} \\
\midrule
\textbf{DCCD} & 16.73 & 53.10 & 70.04 & 66.25 & 51.75 \\
\midrule
Fixed Gate & \cellcolor{negred}$-0.02$ & \cellcolor{negred}$-1.33$ & \cellcolor{posgreen}$+0.22$ & \cellcolor{negred}$-0.18$ & \cellcolor{negred}$-0.78$ \\
Token-Only & \cellcolor{negred}$-6.34$ & \cellcolor{negred}$-1.44$ & \cellcolor{negred}$-0.42$ & \cellcolor{negred}$-0.15$ & \cellcolor{negred}$-0.26$ \\
Doc-Only & \cellcolor{posgreen}$+0.17$ & \cellcolor{negred}$-0.44$ & \cellcolor{posgreen}$+0.07$ & \cellcolor{posgreen}$+0.18$ & \cellcolor{negred}$-0.31$ \\
Random & \cellcolor{negred}$-3.88$ & \cellcolor{negred}$-1.22$ & \cellcolor{negred}$-0.16$ & \cellcolor{posgreen}$+0.11$ & \cellcolor{negred}$-0.21$ \\
\midrule
\multicolumn{6}{c}{@10} \\
\midrule
\textbf{DCCD} & 12.15 & 53.24 & 72.01 & 66.61 & 53.34 \\
\midrule
Fixed Gate & \cellcolor{negred}$-1.06$ & \cellcolor{negred}$-0.33$ & \cellcolor{posgreen}$+0.10$ & \cellcolor{negred}$-0.54$ & \cellcolor{negred}$-0.25$ \\
Token-Only & \cellcolor{negred}$-4.58$ & \cellcolor{negred}$-0.80$ & \cellcolor{negred}$-0.23$ & \cellcolor{negred}$-0.51$ & \cellcolor{negred}$-0.40$ \\
Doc-Only & \cellcolor{negred}$-0.23$ & \cellcolor{negred}$-0.10$ & \cellcolor{negred}$-0.20$ & \cellcolor{negred}$-0.18$ & \cellcolor{negred}$-0.12$ \\
Random & \cellcolor{negred}$-3.70$ & \cellcolor{negred}$-0.72$ & \cellcolor{negred}$-0.05$ & \cellcolor{negred}$-0.25$ & \cellcolor{negred}$-0.34$ \\
\bottomrule
\end{tabular}%
}
\end{center}
\caption{Ablation results with \texttt{Qwen3.5-2B} using the top-5 (@5) and top-10 (@10) documents. Performance differences are reported relative to DCCD, where green indicates improvement and red indicates degradation.}
\label{tab:ablation}
\end{table}

We ablate the main components of \method to understand where the performance gains come from. Specifically, we run \texttt{Qwen3.5-2B} with @5 and @10 on all 5 datasets, and consider the following four settings.
(1) We set gate \(\epsilon_t=1\), removing the adaptive gate during decoding (\textbf{Fixed Gate}). 
(2) We remove the document-level confidence probe, and select contrastive pairs with only token-level confidence (\textbf{Token-Only}).
(3) We remove the token-level confidence probe, and select contrastive pairs with only document-level confidence (\textbf{Doc-Only}).
(4) We randomly select the positive and negative by uniformly sampling from the retrieved documents (\textbf{Random}, with the gate set to the margin of the sampled pair, clipped to $[0,1]$).

The ablation results in \autoref{tab:ablation} show that confidence-based source selection drives the gains of \method, with document-level confidence as the dominant signal. Removing the document-level confidence probe and relying only on token-level confidence produces the largest degradation, especially on DRQA, with drops of $-6.34$ at @5 and $-4.58$ at @10. This suggests that token-level sharpness alone is insufficient in conflict-heavy settings, where a misleading document can still yield a confident local prediction. Randomly selecting the positive and negative documents also hurts performance in most settings, confirming that confidence-based source selection is important for identifying useful contrastive directions.

By contrast, using document-level confidence alone is comparatively stable, and even slightly improves over full \method in three of the five @5 settings. This indicates that document-level answerability is the dominant signal for resolving intra-context conflict, especially on DRQA, where the model must identify the document containing the current ground-truth answer. However, Doc-Only becomes consistently worse at @10 (by 0.10--0.23 points), suggesting that token-level confidence provides a useful complementary signal as the retrieved bundle becomes more distractor-heavy. The fixed-gate ablation is mixed, setting $\epsilon_t=1.0$ slightly improves TQA, but hurts most other settings, including larger drops on NQ and PopQA at @5 and on DRQA at @10. This suggests that always applying the same contrast strength can occasionally help, but the adaptive margin is more robust across datasets and retrieval depths.

\section{Conclusion and Future Work}
\label{sec:conclusion}

In this work, we propose \methodlong (\textbf{\method}), a training-free contrastive decoding method that combines document and token-level confidence to select positive and negative evidence sources during generation, and introduce \textbf{DRQA}, a factual-conflict QA benchmark derived from enterprise deep-research scenarios. Across DRQA and standard RAG benchmarks, \method achieves the best result in 28 of the 30 settings, with the largest gains on DRQA. These results suggest that source-aware, confidence-guided contrast is especially useful when retrieved evidence is internally inconsistent.

For future work, improving the efficiency of \method is an important next step, since the method requires document-conditioned forward passes during decoding. Promising directions include more aggressive caching, document pruning, and lightweight confidence estimation, as well as combining \method with reranking and metadata-aware arbitration.
Second, DRQA currently focuses on single-source factual-conflict QA in enterprise-style settings, extending the benchmark to multi-document synthesis, further conflict types, longer-form answers, richer source metadata, and additional domains would further test retrieval-grounded conflict resolution. We also plan to analyze how performance changes with different combinations and proportions of misinformation, temporal, and noise documents, which are fixed in the current DRQA. Finally, our confidence signals are simple and training-free. Learning or calibrating document-level and token-level confidence estimates may further improve robustness when retrieved evidence is ambiguous or insufficient.

\section*{Limitations}
Similar to other document-aware contrastive decoding methods, \method introduces additional inference cost. At each decoding step, the method requires the full-context logits and document-conditioned logits for the retrieved documents, while the document-level confidence probe is computed once per question--document pair. This cost scales linearly with the number of retrieved documents, which has the same computational complexity as DVD~\citep{jin-etal-2024-dvd}. In practice, our experiments use top-$5$ and top-$10$ retrieval settings, which are common in multi-document QA evaluation, but more efficient implementations such as caching document states, or applying \method only at answer-critical decoding steps could reduce runtime overhead. When a single document is expected to carry the answer, the Doc-Only variant is a constant-cost alternative with three streams per step (Appendix~\ref{app:cost}), and \method can also be applied after a reranking step that reduces the number of documents.

DRQA is designed to isolate retrieval-grounded conflict resolution in enterprise-style multi-document QA, where the answer is unavailable from internal memory and must be recovered from retrieved evidence. This controlled design is useful for evaluating the specific capability targeted by \method, but it is not intended to cover all forms of enterprise RAG. In particular, our current benchmark focuses on factoid question answering over text representations of enterprise documents, rather than full long-form report generation, interactive tool use, or multi-modal file understanding. These settings are closely related to deep-research workflows and could provide valuable extensions of DRQA.
DRQA also assumes exactly one current gold document per question, whereas real bundles may contain several supporting documents or none, and some questions require synthesizing evidence across documents. We do not evaluate such multi-document synthesis. Finally, DRQA covers factual, temporal, and irrelevance distractors, but not authority, scope, partial-context, or interpretation conflicts (\S\ref{sec:drqa-scope}).

Because naturally occurring enterprise conflict corpora are proprietary, DRQA is synthetic. Its correct and noise documents come from human-verified DRBench files, but its misinformation and temporal documents are LLM-generated and filtered by an LLM judge, so their conflicts may differ from those in real enterprise data. The largest gains of \method are on DRQA, while gains on standard QA benchmarks are often below one point. DRQA's document dates also give a staleness cue that a metadata-aware system could use, so our DRQA results measure arbitration without metadata rules.

DRQA accuracy relies on an LLM judge, which we have not validated against human annotations.
 All experiments use small and medium open-weight models (2B--14B).

\section*{Ethical considerations}
DRQA is constructed from synthetic enterprise-style documents and is intended for research on retrieval-grounded conflict resolution, not for representing real companies, employees, or operational records. Because the benchmark includes deliberately generated misinformation and outdated evidence, the dataset should not be used as a factual knowledge source. Its intended use is to evaluate whether RAG systems can identify reliable evidence when retrieved documents conflict. The generated documents use synthetic personas, organizations, and facts. We have verified that the dataset does not contain real private enterprise records or intentionally identifying information. The benchmark is intended for research use only.

\section*{Acknowledgments}
We thank the anonymous reviewers and the area chair for their constructive feedback, which substantially improved this paper.

\bibliography{references/anthology, references/custom, references/camera_ready}

\begin{thebibliography}{45}
\providecommand{\natexlab}[1]{#1}

\bibitem[{Abaskohi et~al.(2026)Abaskohi, Chen, Mu{\~n}oz-M{\'a}rmol, Fox,
  Ramesh, Marcotte, L{\`u}, Chapados, Gella, Pal, Drouin, and
  Laradji}]{abaskohi2026drbench}
Amirhossein Abaskohi, Tianyi Chen, Miguel Mu{\~n}oz-M{\'a}rmol, Curtis Fox,
  Amrutha~Varshini Ramesh, {\'E}tienne Marcotte, Xing~Han L{\`u}, Nicolas
  Chapados, Spandana Gella, Christopher Pal, Alexandre Drouin, and Issam~H.
  Laradji. 2026.
\newblock \href {https://openreview.net/forum?id=IGYQ4c92e2} {{DRB}ench: A
  realistic benchmark for enterprise deep research}.
\newblock In \emph{The Fourteenth International Conference on Learning
  Representations}.

\bibitem[{Abdin et~al.(2024)Abdin, Jacobs, Awan, Aneja, Awadallah,
  Hassan~Awadalla, Bach, Bahree, Bakhtiari, Behl, Benhaim, Bilenko, Bjorck,
  Bubeck, Cai, Mendes, Chen, Chaudhary, Chopra, Giorno, de~Rosa, Dixon, Eldan,
  Iter, Goswami, Gunasekar, Haider, Hao, Hewett, Huynh, Javaheripi, Jin,
  Kauffmann, Karampatziakis, Kim, Khademi, Kurilenko, Lee, Lee, Li, Liang, Liu,
  Lin, Lin, Madan, Mitra, Modi, Nguyen, Norick, Patra, Perez-Becker, Portet,
  Pryzant, Qin, Radmilac, Rosset, Roy, Saarikivi, Saied, Salim, Santacroce,
  Shah, Shang, Sharma, Song, Ruwase, Wang, Ward, Wang, Witte, Wyatt, Xu, Xu,
  Xu, Yadav, Yang, Yang, Yu, Zhang, Zhang, Zhang, Zhang, Zhang, Zhang, and
  Zhou}]{abdin2024phi3}
Marah~I Abdin, Sam~Ade Jacobs, Ammar~Ahmad Awan, Jyoti Aneja, Ahmed Awadallah,
  Hany Hassan~Awadalla, Nguyen Bach, Amit Bahree, Arash Bakhtiari, Harkirat
  Behl, Alon Benhaim, Misha Bilenko, Johan Bjorck, Sébastien Bubeck, Martin
  Cai, Caio César~Teodoro Mendes, Weizhu Chen, Vishrav Chaudhary, Parul
  Chopra, and 66 others. 2024.
\newblock \href
  {https://www.microsoft.com/en-us/research/publication/phi-3-technical-report-a-highly-capable-language-model-locally-on-your-phone/}
  {Phi-3 technical report: A highly capable language model locally on your
  phone}.
\newblock Technical Report MSR-TR-2024-12, Microsoft.

\bibitem[{{Anthropic}(2025)}]{anthropic2025claudeopus45cr}
{Anthropic}. 2025.
\newblock Introducing {Claude Opus 4.5}.
\newblock \url{https://www.anthropic.com/news/claude-opus-4-5}.

\bibitem[{Chen et~al.(2022)Chen, Zhang, and Choi}]{chen-etal-2022-rich}
Hung-Ting Chen, Michael Zhang, and Eunsol Choi. 2022.
\newblock \href {https://doi.org/10.18653/v1/2022.emnlp-main.146} {Rich
  knowledge sources bring complex knowledge conflicts: Recalibrating models to
  reflect conflicting evidence}.
\newblock In \emph{Proceedings of the 2022 Conference on Empirical Methods in
  Natural Language Processing}, pages 2292--2307, Abu Dhabi, United Arab
  Emirates. Association for Computational Linguistics.

\bibitem[{Choubey et~al.(2025)Choubey, Peng, Bhagavath, Huang, Xiong, and
  Wu}]{choubey-etal-2025-benchmarking}
Prafulla~Kumar Choubey, Xiangyu Peng, Shilpa Bhagavath, Kung-Hsiang Huang,
  Caiming Xiong, and Chien-Sheng Wu. 2025.
\newblock \href {https://doi.org/10.18653/v1/2025.emnlp-industry.34}
  {Benchmarking deep search over heterogeneous enterprise data}.
\newblock In \emph{Proceedings of the 2025 Conference on Empirical Methods in
  Natural Language Processing: Industry Track}, pages 501--517, Suzhou (China).
  Association for Computational Linguistics.

\bibitem[{Du et~al.(2026)Du, Xu, Zhu, Zhang, Wang, and
  Mao}]{du2026deepresearch}
Mingxuan Du, Benfeng Xu, Chiwei Zhu, Licheng Zhang, Xiaorui Wang, and Zhendong
  Mao. 2026.
\newblock \href {https://openreview.net/forum?id=hQ0K2Hhq7H} {Deepresearch
  bench: A comprehensive benchmark for deep research agents}.
\newblock In \emph{The Fourteenth International Conference on Learning
  Representations}.

\bibitem[{Fadeeva et~al.(2024)Fadeeva, Rubashevskii, Shelmanov, Petrakov, Li,
  Mubarak, Tsymbalov, Kuzmin, Panchenko, Baldwin, Nakov, and
  Panov}]{fadeeva-etal-2024-fact}
Ekaterina Fadeeva, Aleksandr Rubashevskii, Artem Shelmanov, Sergey Petrakov,
  Haonan Li, Hamdy Mubarak, Evgenii Tsymbalov, Gleb Kuzmin, Alexander
  Panchenko, Timothy Baldwin, Preslav Nakov, and Maxim Panov. 2024.
\newblock \href {https://doi.org/10.18653/v1/2024.findings-acl.558}
  {Fact-checking the output of large language models via token-level
  uncertainty quantification}.
\newblock In \emph{Findings of the Association for Computational Linguistics:
  ACL 2024}, pages 9367--9385, Bangkok, Thailand. Association for Computational
  Linguistics.

\bibitem[{Gao et~al.(2023)Gao, Xiong, Gao, Jia, Pan, Bi, Dai, Sun, Wang, and
  Wang}]{gao2023retrieval}
Yunfan Gao, Yun Xiong, Xinyu Gao, Kangxiang Jia, Jinliu Pan, Yuxi Bi, Yi~Dai,
  Jiawei Sun, Meng Wang, and Haofen Wang. 2023.
\newblock Retrieval-augmented generation for large language models: A survey.
\newblock \emph{arXiv preprint arXiv:2312.10997}.

\bibitem[{Geng et~al.(2024)Geng, Cai, Wang, Koeppl, Nakov, and
  Gurevych}]{geng-etal-2024-survey}
Jiahui Geng, Fengyu Cai, Yuxia Wang, Heinz Koeppl, Preslav Nakov, and Iryna
  Gurevych. 2024.
\newblock \href {https://doi.org/10.18653/v1/2024.naacl-long.366} {A survey of
  confidence estimation and calibration in large language models}.
\newblock In \emph{Proceedings of the 2024 Conference of the North American
  Chapter of the Association for Computational Linguistics: Human Language
  Technologies (Volume 1: Long Papers)}, pages 6577--6595, Mexico City, Mexico.
  Association for Computational Linguistics.

\bibitem[{Hwang et~al.(2025)Hwang, Park, Park, Kim, Park, and
  Ok}]{hwang-etal-2025-retrieval}
Jeongyeon Hwang, Junyoung Park, Hyejin Park, Dongwoo Kim, Sangdon Park, and
  Jungseul Ok. 2025.
\newblock \href {https://doi.org/10.18653/v1/2025.emnlp-main.1738}
  {Retrieval-augmented generation with estimation of source reliability}.
\newblock In \emph{Proceedings of the 2025 Conference on Empirical Methods in
  Natural Language Processing}, pages 34279--34303, Suzhou, China. Association
  for Computational Linguistics.

\bibitem[{Izacard and Grave(2021)}]{izacard-grave-2021-leveraging}
Gautier Izacard and Edouard Grave. 2021.
\newblock \href {https://doi.org/10.18653/v1/2021.eacl-main.74} {Leveraging
  passage retrieval with generative models for open domain question answering}.
\newblock In \emph{Proceedings of the 16th Conference of the European Chapter
  of the Association for Computational Linguistics: Main Volume}, pages
  874--880, Online. Association for Computational Linguistics.

\bibitem[{Java et~al.(2026)Java, Khandelwal, Midigeshi, Halfaker, Deshpande,
  Goyal, Gupta, Natarajan, and Sharma}]{java2026characterizing}
Abhinav Java, Ashmit Khandelwal, Sukruta~Prakash Midigeshi, Aaron Halfaker,
  Amit Deshpande, Navin Goyal, Ankur Gupta, Nagarajan Natarajan, and Amit
  Sharma. 2026.
\newblock \href {https://openreview.net/forum?id=5EmpOCq1Ql} {Characterizing
  deep research: A benchmark and formal definition}.
\newblock In \emph{The Fourteenth International Conference on Learning
  Representations}.

\bibitem[{Jiang et~al.(2023)Jiang, Xu, Gao, Sun, Liu, Dwivedi-Yu, Yang, Callan,
  and Neubig}]{jiang-etal-2023-active}
Zhengbao Jiang, Frank Xu, Luyu Gao, Zhiqing Sun, Qian Liu, Jane Dwivedi-Yu,
  Yiming Yang, Jamie Callan, and Graham Neubig. 2023.
\newblock \href {https://doi.org/10.18653/v1/2023.emnlp-main.495} {Active
  retrieval augmented generation}.
\newblock In \emph{Proceedings of the 2023 Conference on Empirical Methods in
  Natural Language Processing}, pages 7969--7992, Singapore. Association for
  Computational Linguistics.

\bibitem[{Jin et~al.(2024{\natexlab{a}})Jin, Wang, Zhang, Li, and
  Guo}]{jin-etal-2024-dvd}
Jing Jin, Houfeng Wang, Hao Zhang, Xiaoguang Li, and Zhijiang Guo.
  2024{\natexlab{a}}.
\newblock \href {https://doi.org/10.18653/v1/2024.emnlp-main.266} {{DVD}:
  Dynamic contrastive decoding for knowledge amplification in multi-document
  question answering}.
\newblock In \emph{Proceedings of the 2024 Conference on Empirical Methods in
  Natural Language Processing}, pages 4624--4637, Miami, Florida, USA.
  Association for Computational Linguistics.

\bibitem[{Jin et~al.(2024{\natexlab{b}})Jin, Cao, Chen, Liu, Jiang, Xu, Qiuxia,
  and Zhao}]{jin-etal-2024-tug}
Zhuoran Jin, Pengfei Cao, Yubo Chen, Kang Liu, Xiaojian Jiang, Jiexin Xu,
  Li~Qiuxia, and Jun Zhao. 2024{\natexlab{b}}.
\newblock \href {https://aclanthology.org/2024.lrec-main.1466/} {Tug-of-war
  between knowledge: Exploring and resolving knowledge conflicts in
  retrieval-augmented language models}.
\newblock In \emph{Proceedings of the 2024 Joint International Conference on
  Computational Linguistics, Language Resources and Evaluation (LREC-COLING
  2024)}, pages 16867--16878, Torino, Italia. ELRA and ICCL.

\bibitem[{Joren et~al.(2025)Joren, Zhang, Ferng, Juan, Taly, and
  Rashtchian}]{joren2025sufficient}
Hailey Joren, Jianyi Zhang, Chun-Sung Ferng, Da-Cheng Juan, Ankur Taly, and
  Cyrus Rashtchian. 2025.
\newblock Sufficient context: A new lens on retrieval augmented generation
  systems.
\newblock In \emph{International Conference on Learning Representations},
  volume 2025, pages 20310--20334.

\bibitem[{Joshi et~al.(2017)Joshi, Choi, Weld, and
  Zettlemoyer}]{joshi-etal-2017-triviaqa}
Mandar Joshi, Eunsol Choi, Daniel Weld, and Luke Zettlemoyer. 2017.
\newblock \href {https://doi.org/10.18653/v1/P17-1147} {{T}rivia{QA}: A large
  scale distantly supervised challenge dataset for reading comprehension}.
\newblock In \emph{Proceedings of the 55th Annual Meeting of the Association
  for Computational Linguistics (Volume 1: Long Papers)}, pages 1601--1611,
  Vancouver, Canada. Association for Computational Linguistics.

\bibitem[{Kadavath et~al.(2022)Kadavath, Conerly, Askell, Henighan, Drain,
  Perez, Schiefer, Hatfield-Dodds, DasSarma, Tran-Johnson
  et~al.}]{kadavath2022language}
Saurav Kadavath, Tom Conerly, Amanda Askell, Tom Henighan, Dawn Drain, Ethan
  Perez, Nicholas Schiefer, Zac Hatfield-Dodds, Nova DasSarma, Eli
  Tran-Johnson, and 1 others. 2022.
\newblock Language models (mostly) know what they know.
\newblock \emph{arXiv preprint arXiv:2207.05221}.

\bibitem[{Karpukhin et~al.(2020)Karpukhin, Oguz, Min, Lewis, Wu, Edunov, Chen,
  and Yih}]{karpukhin-etal-2020-dense}
Vladimir Karpukhin, Barlas Oguz, Sewon Min, Patrick Lewis, Ledell Wu, Sergey
  Edunov, Danqi Chen, and Wen-tau Yih. 2020.
\newblock \href {https://doi.org/10.18653/v1/2020.emnlp-main.550} {Dense
  passage retrieval for open-domain question answering}.
\newblock In \emph{Proceedings of the 2020 Conference on Empirical Methods in
  Natural Language Processing (EMNLP)}, pages 6769--6781, Online. Association
  for Computational Linguistics.

\bibitem[{Khandelwal et~al.(2025)Khandelwal, Gupta, and
  Agrawal}]{khandelwal-etal-2025-cocoa}
Anant Khandelwal, Manish Gupta, and Puneet Agrawal. 2025.
\newblock \href {https://doi.org/10.18653/v1/2025.emnlp-main.348} {{C}o{C}o{A}:
  Confidence- and context-aware adaptive decoding for resolving knowledge
  conflicts in large language models}.
\newblock In \emph{Proceedings of the 2025 Conference on Empirical Methods in
  Natural Language Processing}, pages 6835--6855, Suzhou, China. Association
  for Computational Linguistics.

\bibitem[{Kim et~al.(2024)Kim, Kim, Park, Park, Cho, Kim, Yoo, Lee, and
  Kim}]{kim-etal-2024-adaptive}
Youna Kim, Hyuhng~Joon Kim, Cheonbok Park, Choonghyun Park, Hyunsoo Cho,
  Junyeob Kim, Kang~Min Yoo, Sang-goo Lee, and Taeuk Kim. 2024.
\newblock \href {https://doi.org/10.18653/v1/2024.findings-emnlp.136} {Adaptive
  contrastive decoding in retrieval-augmented generation for handling noisy
  contexts}.
\newblock In \emph{Findings of the Association for Computational Linguistics:
  EMNLP 2024}, pages 2421--2431, Miami, Florida, USA. Association for
  Computational Linguistics.

\bibitem[{Kudo and Richardson(2018)}]{kudo-richardson-2018-sentencepiece}
Taku Kudo and John Richardson. 2018.
\newblock \href {https://doi.org/10.18653/v1/D18-2012} {{S}entence{P}iece: A
  simple and language independent subword tokenizer and detokenizer for neural
  text processing}.
\newblock In \emph{Proceedings of the 2018 Conference on Empirical Methods in
  Natural Language Processing: System Demonstrations}, pages 66--71, Brussels,
  Belgium. Association for Computational Linguistics.

\bibitem[{Kwiatkowski et~al.(2019)Kwiatkowski, Palomaki, Redfield, Collins,
  Parikh, Alberti, Epstein, Polosukhin, Devlin, Lee, Toutanova, Jones, Kelcey,
  Chang, Dai, Uszkoreit, Le, and Petrov}]{kwiatkowski-etal-2019-natural}
Tom Kwiatkowski, Jennimaria Palomaki, Olivia Redfield, Michael Collins, Ankur
  Parikh, Chris Alberti, Danielle Epstein, Illia Polosukhin, Jacob Devlin,
  Kenton Lee, Kristina Toutanova, Llion Jones, Matthew Kelcey, Ming-Wei Chang,
  Andrew~M. Dai, Jakob Uszkoreit, Quoc Le, and Slav Petrov. 2019.
\newblock \href {https://doi.org/10.1162/tacl_a_00276} {Natural questions: A
  benchmark for question answering research}.
\newblock \emph{Transactions of the Association for Computational Linguistics},
  7:452--466.

\bibitem[{Lee et~al.(2024)Lee, Ye, and Choi}]{leeambigdocs}
Yoonsang Lee, Xi~Ye, and Eunsol Choi. 2024.
\newblock Ambigdocs: Reasoning across documents on different entities under the
  same name.
\newblock In \emph{First Conference on Language Modeling}.

\bibitem[{Lewis et~al.(2020)Lewis, Perez, Piktus, Petroni, Karpukhin, Goyal,
  K\"{u}ttler, Lewis, Yih, Rockt\"{a}schel, Riedel, and
  Kiela}]{lewis2020retrieval}
Patrick Lewis, Ethan Perez, Aleksandra Piktus, Fabio Petroni, Vladimir
  Karpukhin, Naman Goyal, Heinrich K\"{u}ttler, Mike Lewis, Wen-tau Yih, Tim
  Rockt\"{a}schel, Sebastian Riedel, and Douwe Kiela. 2020.
\newblock \href
  {https://proceedings.neurips.cc/paper_files/paper/2020/file/6b493230205f780e1bc26945df7481e5-Paper.pdf}
  {Retrieval-augmented generation for knowledge-intensive nlp tasks}.
\newblock In \emph{Advances in Neural Information Processing Systems},
  volume~33, pages 9459--9474. Curran Associates, Inc.

\bibitem[{Li et~al.(2023)Li, Holtzman, Fried, Liang, Eisner, Hashimoto,
  Zettlemoyer, and Lewis}]{li-etal-2023-contrastive}
Xiang~Lisa Li, Ari Holtzman, Daniel Fried, Percy Liang, Jason Eisner, Tatsunori
  Hashimoto, Luke Zettlemoyer, and Mike Lewis. 2023.
\newblock \href {https://doi.org/10.18653/v1/2023.acl-long.687} {Contrastive
  decoding: Open-ended text generation as optimization}.
\newblock In \emph{Proceedings of the 61st Annual Meeting of the Association
  for Computational Linguistics (Volume 1: Long Papers)}, pages 12286--12312,
  Toronto, Canada. Association for Computational Linguistics.

\bibitem[{Malinin and Gales(2018)}]{malinin-gales-2018-predictive}
Andrey Malinin and Mark Gales. 2018.
\newblock \href
  {https://proceedings.neurips.cc/paper_files/paper/2018/file/3ea2db50e62ceefceaf70a9d9a56a6f4-Paper.pdf}
  {Predictive uncertainty estimation via prior networks}.
\newblock In \emph{Advances in Neural Information Processing Systems},
  volume~31. Curran Associates, Inc.

\bibitem[{Malinin and Gales(2021)}]{malinin2021uncertainty}
Andrey Malinin and Mark Gales. 2021.
\newblock \href {https://openreview.net/forum?id=jN5y-zb5Q7m} {Uncertainty
  estimation in autoregressive structured prediction}.
\newblock In \emph{International Conference on Learning Representations}.

\bibitem[{Mallen et~al.(2023)Mallen, Asai, Zhong, Das, Khashabi, and
  Hajishirzi}]{mallen-etal-2023-trust}
Alex Mallen, Akari Asai, Victor Zhong, Rajarshi Das, Daniel Khashabi, and
  Hannaneh Hajishirzi. 2023.
\newblock \href {https://doi.org/10.18653/v1/2023.acl-long.546} {When not to
  trust language models: Investigating effectiveness of parametric and
  non-parametric memories}.
\newblock In \emph{Proceedings of the 61st Annual Meeting of the Association
  for Computational Linguistics (Volume 1: Long Papers)}, pages 9802--9822,
  Toronto, Canada. Association for Computational Linguistics.

\bibitem[{Nguyen et~al.(2026)Nguyen, Gupta, and Le}]{nguyen2026probabilities}
Manh Nguyen, Sunil Gupta, and Hung Le. 2026.
\newblock \href {https://doi.org/10.1609/aaai.v40i38.40531} {Probabilities are
  all you need: A probability-only approach to uncertainty estimation in large
  language models}.
\newblock \emph{Proceedings of the AAAI Conference on Artificial Intelligence},
  40(38):32546–32554.

\bibitem[{{Qwen Team}(2026)}]{qwen35blogcr}
{Qwen Team}. 2026.
\newblock \href {https://qwen.ai/blog?id=qwen3.5} {{Qwen3.5}: Accelerating
  productivity with native multimodal agents}.

\bibitem[{Schuster et~al.(2026)Schuster, Gautam, and
  Markert}]{schuster2026whose}
Jakob Schuster, Vagrant Gautam, and Katja Markert. 2026.
\newblock Whose facts win? {LLM} source preferences under knowledge conflicts.
\newblock \emph{arXiv preprint arXiv:2601.03746}.

\bibitem[{Sennrich et~al.(2016)Sennrich, Haddow, and
  Birch}]{sennrich-etal-2016-neural}
Rico Sennrich, Barry Haddow, and Alexandra Birch. 2016.
\newblock \href {https://doi.org/10.18653/v1/P16-1162} {Neural machine
  translation of rare words with subword units}.
\newblock In \emph{Proceedings of the 54th Annual Meeting of the Association
  for Computational Linguistics (Volume 1: Long Papers)}, pages 1715--1725,
  Berlin, Germany. Association for Computational Linguistics.

\bibitem[{Shi et~al.(2024)Shi, Han, Lewis, Tsvetkov, Zettlemoyer, and
  Yih}]{shi-etal-2024-trusting}
Weijia Shi, Xiaochuang Han, Mike Lewis, Yulia Tsvetkov, Luke Zettlemoyer, and
  Wen-tau Yih. 2024.
\newblock \href {https://doi.org/10.18653/v1/2024.naacl-short.69} {Trusting
  your evidence: Hallucinate less with context-aware decoding}.
\newblock In \emph{Proceedings of the 2024 Conference of the North American
  Chapter of the Association for Computational Linguistics: Human Language
  Technologies (Volume 2: Short Papers)}, pages 783--791, Mexico City, Mexico.
  Association for Computational Linguistics.

\bibitem[{Singh et~al.(2023)Singh, Morris, Rush, Gao, and
  Deng}]{singh-etal-2023-tree}
Chandan Singh, John Morris, Alexander Rush, Jianfeng Gao, and Yuntian Deng.
  2023.
\newblock \href {https://doi.org/10.18653/v1/2023.emnlp-main.384} {Tree
  prompting: Efficient task adaptation without fine-tuning}.
\newblock In \emph{Proceedings of the 2023 Conference on Empirical Methods in
  Natural Language Processing}, pages 6253--6267, Singapore. Association for
  Computational Linguistics.

\bibitem[{Wan et~al.(2024)Wan, Wallace, and Klein}]{wan-etal-2024-evidence}
Alexander Wan, Eric Wallace, and Dan Klein. 2024.
\newblock \href {https://doi.org/10.18653/v1/2024.acl-long.403} {What evidence
  do language models find convincing?}
\newblock In \emph{Proceedings of the 62nd Annual Meeting of the Association
  for Computational Linguistics (Volume 1: Long Papers)}, pages 7468--7484,
  Bangkok, Thailand. Association for Computational Linguistics.

\bibitem[{Wan et~al.(2025)Wan, Wang, Li, Liu, Zhu, and
  Zhu}]{wan2025pokeeresearch}
Yi~Wan, Jiuqi Wang, Liam Li, Jinsong Liu, Ruihao Zhu, and Zheqing Zhu. 2025.
\newblock Rethinking the design of reinforcement learning-based deep research
  agents.
\newblock \emph{arXiv preprint arXiv:2510.15862}.

\bibitem[{Wang et~al.(2025{\natexlab{a}})Wang, Prasad, Stengel-Eskin, and
  Bansal}]{wang-etal-2025-adacad}
Han Wang, Archiki Prasad, Elias Stengel-Eskin, and Mohit Bansal.
  2025{\natexlab{a}}.
\newblock \href {https://doi.org/10.18653/v1/2025.naacl-long.581} {{A}da{CAD}:
  Adaptively decoding to balance conflicts between contextual and parametric
  knowledge}.
\newblock In \emph{Proceedings of the 2025 Conference of the Nations of the
  Americas Chapter of the Association for Computational Linguistics: Human
  Language Technologies (Volume 1: Long Papers)}, pages 11636--11652,
  Albuquerque, New Mexico. Association for Computational Linguistics.

\bibitem[{Wang et~al.(2025{\natexlab{b}})Wang, Prasad, Stengel-Eskin, and
  Bansal}]{wangretrieval}
Han Wang, Archiki Prasad, Elias Stengel-Eskin, and Mohit Bansal.
  2025{\natexlab{b}}.
\newblock Retrieval-augmented generation with conflicting evidence.
\newblock In \emph{Second Conference on Language Modeling}.

\bibitem[{Wolf et~al.(2020)Wolf, Debut, Sanh, Chaumond, Delangue, Moi, Cistac,
  Rault, Louf, Funtowicz, Davison, Shleifer, von Platen, Ma, Jernite, Plu, Xu,
  Le~Scao, Gugger, Drame, Lhoest, and Rush}]{wolf-etal-2020-transformers}
Thomas Wolf, Lysandre Debut, Victor Sanh, Julien Chaumond, Clement Delangue,
  Anthony Moi, Pierric Cistac, Tim Rault, Remi Louf, Morgan Funtowicz, Joe
  Davison, Sam Shleifer, Patrick von Platen, Clara Ma, Yacine Jernite, Julien
  Plu, Canwen Xu, Teven Le~Scao, Sylvain Gugger, and 3 others. 2020.
\newblock \href {https://doi.org/10.18653/v1/2020.emnlp-demos.6} {Transformers:
  State-of-the-art natural language processing}.
\newblock In \emph{Proceedings of the 2020 Conference on Empirical Methods in
  Natural Language Processing: System Demonstrations}, pages 38--45, Online.
  Association for Computational Linguistics.

\bibitem[{Xie et~al.(2024)Xie, Zhang, Chen, Lou, and Su}]{xie2024adaptive}
Jian Xie, Kai Zhang, Jiangjie Chen, Renze Lou, and Yu~Su. 2024.
\newblock \href {https://openreview.net/forum?id=auKAUJZMO6} {Adaptive
  chameleon or stubborn sloth: Revealing the behavior of large language models
  in knowledge conflicts}.
\newblock In \emph{The Twelfth International Conference on Learning
  Representations}.

\bibitem[{Xu et~al.(2024)Xu, Qi, Guo, Wang, Wang, Zhang, and
  Xu}]{xu-etal-2024-knowledge-conflicts}
Rongwu Xu, Zehan Qi, Zhijiang Guo, Cunxiang Wang, Hongru Wang, Yue Zhang, and
  Wei Xu. 2024.
\newblock \href {https://doi.org/10.18653/v1/2024.emnlp-main.486} {Knowledge
  conflicts for {LLM}s: A survey}.
\newblock In \emph{Proceedings of the 2024 Conference on Empirical Methods in
  Natural Language Processing}, pages 8541--8565, Miami, Florida, USA.
  Association for Computational Linguistics.

\bibitem[{Zhang et~al.(2024)Zhang, Fang, and
  Chen}]{zhang-etal-2024-retrievalqa}
Zihan Zhang, Meng Fang, and Ling Chen. 2024.
\newblock \href {https://doi.org/10.18653/v1/2024.findings-acl.415}
  {{R}etrieval{QA}: Assessing adaptive retrieval-augmented generation for
  short-form open-domain question answering}.
\newblock In \emph{Findings of the Association for Computational Linguistics:
  ACL 2024}, pages 6963--6975, Bangkok, Thailand. Association for Computational
  Linguistics.

\bibitem[{Zhao et~al.(2024)Zhao, Monti, Lehmann, and
  Assem}]{zhao-etal-2024-enhancing}
Zheng Zhao, Emilio Monti, Jens Lehmann, and Haytham Assem. 2024.
\newblock \href {https://doi.org/10.18653/v1/2024.naacl-long.237} {Enhancing
  contextual understanding in large language models through contrastive
  decoding}.
\newblock In \emph{Proceedings of the 2024 Conference of the North American
  Chapter of the Association for Computational Linguistics: Human Language
  Technologies (Volume 1: Long Papers)}, pages 4225--4237, Mexico City, Mexico.
  Association for Computational Linguistics.

\bibitem[{Zheng et~al.(2025)Zheng, Fu, Hu, Cai, Ye, Lu, and
  Liu}]{zheng-etal-2025-deepresearcher}
Yuxiang Zheng, Dayuan Fu, Xiangkun Hu, Xiaojie Cai, Lyumanshan Ye, Pengrui Lu,
  and Pengfei Liu. 2025.
\newblock \href {https://doi.org/10.18653/v1/2025.emnlp-main.22}
  {{D}eep{R}esearcher: Scaling deep research via reinforcement learning in
  real-world environments}.
\newblock In \emph{Proceedings of the 2025 Conference on Empirical Methods in
  Natural Language Processing}, pages 414--431, Suzhou, China. Association for
  Computational Linguistics.

\end{thebibliography}

\appendix

\section{DRQA Construction Details}
\label{app:drqa-details}
This section describes the procedure used to construct DRQA from DRBench.
For each selected DRBench internal insight, we construct one DRQA example. Each example consists of a question, a ground-truth answer, and a retrieved document bundle. The question is the specific question that DRBench provides for the internal insight (\texttt{qa\_dict.json}) and asks for a specific enterprise-contextualized value, such as a percentage, dollar amount, date, count, or named entity. The ground-truth answer is the original DRBench internal insight. For example, an insight such as ``Lee's Market reduced food waste by 8\% in Q2 2024, saving \$1.2M'' comes with the question ``What is Lee's Market's current food waste reduction rate as of Q2 2024?''.

\subsection{DRBench Background}
\label{app:drbench-background}

DRBench~\citep{abaskohi2026drbench} is an enterprise deep-research benchmark designed to evaluate agents on complex, open-ended research tasks. Unlike standard question answering benchmarks \citep{kwiatkowski-etal-2019-natural, izacard-grave-2021-leveraging, mallen-etal-2023-trust, zhang-etal-2024-retrievalqa}, which usually ask for a short factual answer from a public corpus, DRBench tasks require an agent to synthesize information across public and private enterprise sources and produce a structured research report. Each task is grounded in an enterprise scenario that specifies a company context, a user persona, and a broad deep-research question. The retrieval environment includes heterogeneous sources such as public web pages, productivity documents, cloud files, emails, and chat conversations.

Each DRBench task contains supporting facts, referred to as \emph{insights}, that are relevant to answering the deep-research question. These insights include both external facts that may be available on the public web and internal facts that are generated as private enterprise knowledge. The internal facts are embedded in generated enterprise documents using a ``needle-in-a-haystack construction''. Specifically, a realistic document is first outlined in a target modality, the insight is inserted into an appropriate section, and the remaining content is filled with contextually plausible but topically irrelevant text. This construction creates realistic enterprise files in which the answer-bearing evidence is embedded among surrounding noise.

DRQA reuses DRBench as scenario scaffolding rather than as a long-form report-generation benchmark. In particular, we take the synthetically generated internal insights from DRBench, which are not expected to be recoverable from the model's internal memory, and convert them into short-answer multi-document question answering tasks. This allows us to isolate the intermediate RAG capability we study in this paper: given a question and a bundle of retrieved documents, the model must identify which document contains the most reliable evidence and commit to the correct answer.

\subsection{Document Generation Prompts}
\label{app:gen-prompts}

We list the prompt templates used during DRQA construction. Each template is shown as a Python f-string as passed to the LLM. Unless otherwise stated, prompts were run with \texttt{anthropic/claude-opus-4-5} \cite{anthropic2025claudeopus45cr} through OpenRouter\footnote{\url{https://openrouter.ai}}.

\paragraph{Wrong-answer generation for misinformation documents}
This prompt produces the perturbed value inserted into each misinformation document. It is called once per misinformation document before the format-specific document generator. The document date is passed to the prompt so that dates or time periods in the wrong answer fall after the document date.

\begin{Prompt}
Given this correct answer, generate a WRONG but plausible alternative.

CORRECT ANSWER: {fact.answer}
QUESTION: {fact.question}

RULES:
- If the answer contains a number/percentage, change it significantly (e.g., 8% -> 15% or 3%)
- If it contains a date, shift it by 6-18 months in either direction
- The document containing this answer is dated {doc_date}. Any dates or time periods in the wrong answer MUST be after this document date (i.e. cannot reference events that have already passed relative to the document date).
- Keep the same format/type as the original
- Make it plausible but clearly different

Return JSON:
{"wrong_answer": "The wrong answer", "reason": "Brief explanation of what was changed"}
\end{Prompt}

\paragraph{Historical-answer generation for temporal documents}
This prompt produces the historical value inserted into a temporal document. The document date is sampled first and passed to the LLM, ensuring that the generated value is coherent with the document's timestamp.

\begin{Prompt}
You are writing a fact for a document dated {doc_date}.
The document will present this fact as its CURRENT finding -- not as
a historical reference.

CURRENT (most recent) answer: {fact.answer}
QUESTION: {fact.question}

The document that will contain this answer is dated {doc_date}.
Generate a value that would have been accurate and considered
current as of that date.

STRICT RULES:
- The time period in your answer MUST differ from the time period
  in the current answer.
  e.g., if current says "Q2 2024", your answer must say "Q1 2023" or
  "Q4 2022", NOT "Q2 2024".
- Change the numeric value OR the referenced date/period -- both if
  possible.
- Do NOT use the phrase "as of [same date as current answer]".
- The value can be higher or lower than current -- both directions
  are valid.

Return JSON:
{"temporal_answer": "The value as it stood on {doc_date}",
 "reason": "Brief note on what period this covers"}
\end{Prompt}

\paragraph{Additional context for misinformation documents}
For misinformation documents, we append a format-specific uncertainty context to the document-generation prompt. This context encourages the generated document to express the incorrect claim with subtle unreliability markers rather than obvious disclaimers. Two phrases are sampled uniformly without replacement from the pool corresponding to the document format, shown in \autoref{tab:uncertainty-phrases}.

\begin{Prompt}
Document date: {formatted_date}. This contains INCORRECT information
with subtle uncertainty markers. {phrase_1} {phrase_2} Keep
uncertainty natural and subtle, not obvious.
\end{Prompt}

\begin{table*}[t]
\centering
\small
\begin{tabularx}{\linewidth}{@{}lX@{}}
\toprule
\textbf{Format} & \textbf{Phrase pool; two sampled per misinformation document} \\
\midrule
email &
Use casual hedging such as ``I think'' or ``from what I understand''; mention hearing the information from a colleague or in passing; use phrases such as ``if I remember correctly'' or ``I believe''; frame the message as informal knowledge sharing rather than official communication; include a casual tone suggesting secondhand information. \\
\midrule
chat &
Use casual language such as ``pretty sure'' or ``I think so''; frame the claim as something heard from someone else; include conversational uncertainty such as ``might be'' or ``could be''; use informal phrasing that suggests incomplete knowledge; add hedging such as ``afaik'' or ``last I heard''. \\
\midrule
md &
Include qualifying language such as ``preliminary findings'' or ``initial assessment''; use phrases such as ``based on available information'' or ``current understanding''; frame conclusions tentatively with ``appears to'' or ``suggests that''; include a note that information is subject to verification; use conditional language such as ``may'' or ``could'' for key claims. \\
\bottomrule
\end{tabularx}
\caption{Format-specific uncertainty phrase pools for misinformation documents. Two phrases are sampled uniformly without replacement for each generated misinformation document.}
\label{tab:uncertainty-phrases}
\end{table*}

\paragraph{Additional context for temporal documents}
For temporal documents, we append a format-specific context block that makes the document read as authoritative at the time of writing. The document date is the only staleness signal.
Two phrases are sampled uniformly from the format-specific pool in \autoref{tab:temporal-phrases}.

\begin{Prompt}
Document date: {formatted_date}. Write this as the current state of
affairs as of the document date above -- the author believes this
information is accurate right now. Do NOT reference it as historical,
archived, or outdated. Do NOT use phrases like 'back in',
'previously', or 'as of [earlier date]'. The date field is the only
staleness signal; the content must read as authoritative for its
time. {phrase_1} {phrase_2}
\end{Prompt}

\begin{table*}[t]
\centering
\small
\begin{tabularx}{\linewidth}{@{}lX@{}}
\toprule
\textbf{Format} & \textbf{Phrase pool; two sampled per temporal document} \\
\midrule
md &
Write the document as if it were current at the time of writing; use present tense to describe the information as valid; make the document read as authoritative for its time period; include natural date references that match the document date. \\
\midrule
email &
Write the email naturally as if sent on that date; have the sender convey the information confidently; use normal business communication style for the time; include a natural email signature with the date context. \\
\midrule
chat &
Write the chat as a natural conversation from that time; have participants discuss the topic as current information; use casual workplace-chat language; include natural timestamp references. \\
\bottomrule
\end{tabularx}
\caption{Format-specific context phrases for temporal documents. Two phrases are sampled uniformly per temporal document.}
\label{tab:temporal-phrases}
\end{table*}

\paragraph{Format-specific document generators}
After the wrong-answer or temporal-answer prompt produces a perturbed answer, we call a format-specific generator to produce the final document. The default \texttt{word\_limit} values are 300 for email, 300 for chat, and 500 for Markdown. JSONL outputs are parsed line by line. Markdown outputs are stripped of surrounding Markdown code fences when present.
\vspace{1em}

\noindent Topic Variable Prompt:
\begin{Prompt}
Topic: {category}
Insight: {question} {answer}
Question: {question}
Answer: {answer}
Context: {additional_context}
\end{Prompt}

\noindent Email Generator Prompt:
\begin{Prompt}
You are an email expert who creates professional emails with proper
structure and formatting.
Create a JSONL file that contains a series of email messages in the
mailboxes of persona: {persona.first_name} {persona.last_name},
role: {persona.role}, department: {persona.department},
responsibilities: {persona.responsibilities}.
The emails need to cover the following information:

{topic}

The emails should:
1. Be realistic email conversations that mention the answer naturally to the question 
2. Follow this exact format for each email:
   {
       "type": "email",
       "id": "email_001",
       "from": "sender@company.com",
       "from_name": "Sender Name",
       "to": ["recipient@company.com"],
       "cc": [],
       "subject": "Email Subject",
       "date": "2025-01-20T09:00:00-05:00",
       "body": "Email body content...",
       "folder": "inbox",
       "read": false,
       "attachments": []
   }
3. Have the persona as the sender or one of the recipients
4. Be approximately {word_limit} words in total across all emails
5. Use realistic names, email addresses, and subjects based on the topic
6. Use sequential email IDs
7. Use realistic recent dates
8. Use folders from: inbox, sent, drafts, spam, trash
9. Mark some emails as read and some as unread

Consider generating:
- Emails between team members discussing the topic
- Emails from management asking for specific information
- Internal announcements or updates

Wrap the email content in a JSONL code block.

Company Context:
- {company_info_key_1}: {company_info_value_1}
- ...

Return only the complete email content in JSONL format.
\end{Prompt}

\noindent Chat Generator Prompt
\begin{Prompt}
You are a MatterMost chat expert who creates professional chats with
proper structure, tables, and formatting. Create a JSONL file that
contains a series of chat messages in channels. The messages need to
cover the following information:

{topic}

The chat should:
1. Be a discussion between reasonable users on the topic and mention
   the answer to the question inherently
2. Contain realistic messages to teams/channels
3. Contain all necessary team, channel, and user settings before posts
4. Be approximately {word_limit} words in total
5. Use realistic names for people, teams, and channels
6. Contain a version object at the beginning of the file
7. Order objects as: version, teams, channels, users, posts
8. Include user credential information and team membership

Consider generating:
- Chats between customers/partners and support/sales/product teams
- Chats between team members
- Chats between employees and management

Wrap the chat content in a JSONL code block.

Company Context:
- {company_info_key_1}: {company_info_value_1}
- ...

Return only the complete chat content in JSONL format.
\end{Prompt}

\noindent Markdown Document Generator Prompt:
\begin{Prompt}
You are a markdown expert who creates professional documents with
proper structure, tables, and formatting. Create a complete markdown
document given the following requirements.

{topic}

The document should:
1. Use proper markdown headings
2. Be approximately {word_limit} words in total
3. Include at least one table with relevant data
4. Use proper markdown formatting for lists, code blocks, blockquotes,
   and links where appropriate
5. Use professional terminology and relevant examples
6. Use realistic names for people
7. Use a random name for the document author
8. Include a title and author at the beginning
9. Never include file paths

Company Context:
- {company_info_key_1}: {company_info_value_1}
- ...

Return only the complete markdown content.
\end{Prompt}

\subsection{LLM-as-Judge Verification Prompts}
Each misinformation, temporal, and noise document (noise documents are reused DRBench distractors) is independently checked by an LLM-as-judge using a fixed type-specific rubric. The judge returns structured JSON containing a binary verdict, a one-sentence rationale, and a scalar confidence in $[0,1]$. The judge receives the question, the ground-truth answer, the claimed answer when applicable, the document date when applicable, and the first 2000 characters of the document text. Correct documents copied from DRBench source files are auto-passed because they were already validated in DRBench; applying a 2000-character truncation to long source documents can otherwise create false negatives when the answer appears later in the file.

\paragraph{Correct-Document Rubric Prompt:}
\begin{Prompt}
You are quality-checking a "correct" document in a factual QA
conflict benchmark.

Question: {question}
Ground-truth answer: {gt_answer}
Document excerpt:
{text}

Does this document explicitly state or clearly entail the ground-truth answer?

Respond with valid JSON only:
{
  "verdict": "PASS" or "FAIL",
  "reason": "<one sentence>",
  "confidence": <float 0.0-1.0>
}

PASS: the document contains the ground-truth answer verbatim or as an unambiguous paraphrase.
FAIL: the ground-truth answer cannot be inferred from this document alone.
\end{Prompt}

\paragraph{Misinformation-Document Rubric Prompt:}
\begin{Prompt}
You are quality-checking a "misinformation" document in a factual
QA conflict benchmark.

Question: {question}
Ground-truth answer (correct): {gt_answer}
Claimed answer in this document: {claimed_answer}
Document excerpt:
{text}

Verify two properties:
(A) DISTINCT -- the claimed answer is factually and specifically different from the ground-truth answer.
(B) PLAUSIBLE -- the claimed answer sounds credible to a naive reader and is not obviously absurd.

Respond with valid JSON only:
{
  "verdict": "PASS" or "FAIL",
  "distinct": <true/false>,
  "plausible": <true/false>,
  "reason": "<one sentence>",
  "confidence": <float 0.0-1.0>
}

PASS requires distinct=true AND plausible=true.
\end{Prompt}

\paragraph{Temporal-Document Rubric Prompt:}
\label{app:verify-temporal}

\begin{Prompt}
You are quality-checking a "temporal conflict" document in a factual
QA conflict benchmark.

Question: {question}
Current ground-truth answer: {gt_answer}
Historical answer in this document: {claimed_answer}
Document date: {document_date}
Document excerpt:
{text}

Verify two properties:
(A) DISTINCT -- the historical answer differs meaningfully from the
    current ground-truth answer (different number, date, or factual
    claim). Do NOT judge whether the direction of change is plausible --
    metrics can improve or worsen over time for many valid business
    reasons.
(B) NOT_LABELLED -- the document does NOT contain explicit "archived",
    "outdated", or "superseded" labels. The document date is the only
    staleness signal; the text should read as authoritative for its time.

Respond with valid JSON only:
{
  "verdict": "PASS" or "FAIL",
  "distinct": <true/false>,
  "not_labelled": <true/false>,
  "reason": "<one sentence>",
  "confidence": <float 0.0-1.0>
}

PASS requires distinct=true AND not_labelled=true.
\end{Prompt}

\paragraph{Noise-Document Rubric Prompt:}
\label{app:verify-noise}

\begin{Prompt}
You are quality-checking a "noise" document in a factual QA conflict
benchmark.

Question: {question}
Ground-Truth answer (for reference): {gt_answer}
Document excerpt:
{text}

Does this document NOT answer the target question? That is, is it
genuinely about a different topic and does not contain or clearly imply the ground-truth answer?

Respond with valid JSON only:
{
  "verdict": "PASS" or "FAIL",
  "reason": "<one sentence>",
  "confidence": <float 0.0-1.0>
}

FAIL if the document contains the ground-truth answer or makes a strongly similar claim.
PASS if the document is about a different topic entirely.
\end{Prompt}

\subsection{PPTX Markdown Reconstruction}
\label{app:pptx-reconstruction}
Since DRBench PPTX files do not include an exported text representation,  we avoid using an LLM for deterministic reformatting and render the structured \texttt{file\_dict.json} metadata as Markdown. PDF and DOCX files are read from their DRBench Markdown exports, and XLSX files from their CSV exports. Email and chat files keep DRBench's JSONL format, whose first line holds the persona's synthetic mailbox account record.

\begin{Prompt}
def render_pptx_as_markdown(fd: dict) -> str:
    lines = [f"# {fd.get('file_title', 'Presentation')}", ""]
    intro = fd.get("introduction", "").strip()
    if intro:
        lines += ["## Introduction", "", intro, ""]
    for section in fd.get("subsections", []):
        heading = section.get("heading", "").strip()
        content = section.get("content", "").strip()
        if heading:
            lines += [f"## {heading}", ""]
        if content:
            lines += [content, ""]
    conclusion = fd.get("conclusion", "").strip()
    if conclusion:
        lines += ["## Conclusion", "", conclusion, ""]
    return "\n".join(lines)
\end{Prompt}

\noindent This reconstruction preserves the text fields but does not attempt to preserve slide layout or visual formatting.

\subsection{Final Dataset}
\label{app:bundle-stats}

Table~\ref{tab:bundle-config} summarizes the two released DRQA bundle configurations. Both configurations are derived from the same source examples. The @10 configuration adds one additional misinformation document, one additional temporal document, and three additional noise documents relative to @5. Document dates are stored as metadata and are not part of the model input. The body of a generated document often states its own date (e.g., in an email header or a chat timestamp). In the @10 bundles, this holds for at least 786 of the 1,136 temporal and 441 of the 1,136 misinformation documents, but for none of the 568 correct documents.

\begin{table}[ht]
\centering
\small
\setlength{\tabcolsep}{4pt}
\begin{tabular}{@{}lcc@{}}
\toprule
& \textbf{@5} & \textbf{@10} \\
\midrule
Correct        & 1 & 1 \\
Misinformation & 1 & 2 \\
Temporal       & 1 & 2 \\
Noise          & 2 & 5 \\
\midrule
\textbf{Total documents per example} & \textbf{5} & \textbf{10} \\
\textbf{Examples in release} & 568 & 568 \\
\textbf{Examples at exact-$k$ bundle size} & 568 & 564 \\
\bottomrule
\end{tabular}
\caption{DRQA bundle configurations. All 568 examples in both configurations have the listed numbers of correct, misinformation, and temporal documents. For @10, four examples have fewer than five noise documents because the corresponding DRBench tasks have insufficient distinct distractors after deduplication or repeated noise-rubric rejection.}
\label{tab:bundle-config}
\end{table}

\subsection[Conflict Taxonomy]{Conflict Taxonomy}
\label{app:conflict-taxonomy}
\autoref{tab:conflict-taxonomy} places the DRQA document types in a broader taxonomy of conflicts that arise in enterprise retrieval. DRQA covers factual, temporal, and irrelevance distractors. Its misinformation documents carry hedges typical of informal communication, which is a weak authority cue, but DRQA does not model conflicts between sources of different standing. The remaining types require annotations or metadata that DRBench does not provide, such as approval status or organizational scope, and we leave them to future work. \method does not model any of these types explicitly.

\begin{table*}[t]
\centering
\small
\begin{tabularx}{\linewidth}{@{}lXc@{}}
\toprule
\textbf{Conflict type} & \textbf{Description and enterprise example} & \textbf{In DRQA} \\
\midrule
Factual (misinformation) & A document states a plausible but incorrect value, e.g., a chat message that misremembers a quarterly figure. & \checkmark \\
Temporal & A document states a value that was correct at an earlier date, e.g., last year's figure presented as current. The document date is the only cue. & \checkmark \\
Irrelevance (noise) & A topically related document that does not answer the question. & \checkmark \\
\midrule
Authority & Sources of different standing disagree, e.g., a draft slide deck versus a finance-approved report. & -- \\
Scope mismatch & Values refer to different scopes, e.g., a regional versus a company-wide figure. & -- \\
Partial context & A document supports only part of the answer, e.g., the value but not the period it refers to. & -- \\
Interpretation & The same data are summarized differently, e.g., one team reports an improvement and another a decline. & -- \\
\bottomrule
\end{tabularx}
\caption{Conflict types in enterprise retrieval and their coverage in DRQA.}
\label{tab:conflict-taxonomy}
\end{table*}

\section{DCCD Implementation Details}
\label{app:dccd}

This appendix provides implementation details for \methodlong~(\textbf{\method}), complementing the method description in \S\ref{sec:method}. 

\subsection{Support Probe for Document-Level Confidence}
\label{app:support-probe}

For each question--document pair $(x,d_i)$, we compute a document-level confidence score $q_i$ before decoding. The goal of $q_i$ is to estimate whether document $d_i$ contains enough information to answer the question, independent of the current generation prefix.

\paragraph{Prompt Template}
For each retrieved document $d_i$, we use the following confidence probe:

\begin{Prompt}
{document}

Question: {question}
Does this document contain enough information to answer the question? (yes/no):
Answer:
\end{Prompt}

\paragraph{Yes/No Verbalizers}
Let $\mathcal{Y}_{+}$ denote the set of yes verbalizer token ids and $\mathcal{Y}_{-}$ denote the set of no verbalizer token ids. In practice, we construct these sets by tokenizing common case and leading-space variants:
\[
\{\texttt{yes}, \texttt{Yes}, \texttt{YES}, \texttt{ yes}, \texttt{ Yes}, \texttt{ YES}\}
\]
for yes, and analogously for no. We keep only variants that are represented as a single token by the tokenizer. Leading-space variants are included because the BPE \citep{sennrich-etal-2016-neural} for \texttt{Qwen3.5} and the SentencePiece tokenizers \citep{kudo-richardson-2018-sentencepiece} encode the first content token of an assistant response with an initial whitespace marker.

\paragraph{Score Computation}
We compute the document-level confidence score by re-normalizing the probability mass assigned to yes and no verbalizers. In practice, we compute the same quantity in logit space for numerical stability. Let $\mathbf{z}^{\mathrm{sup}}_i$ be the final-position logits produced by the support probe for document $d_i$. We compute
\begin{align}
\ell^+_i
&=
\operatorname{logsumexp}_{v\in \mathcal{Y}_{+}}
\mathbf{z}^{\mathrm{sup}}_i[v],
\\
\ell^-_i
&=
\operatorname{logsumexp}_{v\in \mathcal{Y}_{-}}
\mathbf{z}^{\mathrm{sup}}_i[v],
\\
q_i
&=
\sigma(\ell^+_i-\ell^-_i),
\label{eq:app-q-logspace}
\end{align}
where $\sigma$ is the logistic sigmoid. This is equivalent to the yes/no renormalization in Eq.~\ref{eq:q-answerability}. All document probes for a query are batched into a single forward pass, so the support-probe stage is computed once before decoding.

\subsection{Token-Level Confidence}
\label{app:token-confidence}

At decoding step \(t\), let \(\mathbf{z}^i_t\in\mathbb{R}^{V}\) denote the next-token logits conditioned on the question \(x\), document \(d_i\), and the current output prefix \(y_{<t}\). Let \(\mathcal{T}_{i,t}\) be the set of top-\(k\) token indices under \(\mathbf{z}^i_t\). We compute token-level confidence by first mapping these top-\(k\) logits to positive evidence values and then measuring the expected entropy of the induced distribution.

\paragraph{Top-\(k\) Evidence}
Following the Dirichlet uncertainty framework of \citet{malinin-gales-2018-predictive}, we convert each selected logit into a positive concentration parameter for a lightweight Dirichlet approximation:
\begin{equation}
\alpha_{(i,t,v)}
=
\operatorname{softplus}\!\left(\mathbf{z}^i_t[v]\right)+1,
\qquad
v\in\mathcal{T}_{(i,t)}.
\label{eq:app-alpha}
\end{equation}
The softplus transformation keeps the evidence positive without exponentiating logits, and the \(+1\) shift ensures that each concentration parameter is at least one. We define the total concentration and the corresponding Dirichlet mean as
\begin{equation}
S_{(i,t)}
=
\sum_{v\in\mathcal{T}_{i,t}}
\alpha_{(i,t,v)},
\qquad
\bar p_{(i,t,v)}
=
\frac{\alpha_{(i,t,v)}}{S_{(i,t)}}.
\label{eq:app-dirichlet-mean}
\end{equation}

\paragraph{Expected Entropy}
We take the top-\(k\) next-token probabilities \citep{nguyen2026probabilities} and treat them as a categorical distribution sampled from a Dirichlet posterior:
\begin{equation}
\boldsymbol{\pi}^i_t
\sim
\operatorname{Dir}\!\left(\boldsymbol{\alpha}^i_t\right).
\end{equation}
The token-level uncertainty is the expected entropy of this sampled categorical distribution:
\begin{equation}
H_k(\mathbf{z}^i_t)
=
\mathbb{E}_{\boldsymbol{\pi}^i_t\sim
\operatorname{Dir}(\boldsymbol{\alpha}^i_t)}
\left[
H(\boldsymbol{\pi}^i_t)
\right].
\label{eq:app-raw-entropy-def}
\end{equation}%
This expectation has the closed form
\begin{equation}
\begin{split}
H_k(\mathbf{z}^i_t)
={}&
\sum_{v\in\mathcal{T}_{(i,t)}}
\bar p_{(i,t,v)}
\bigl[
\psi(S_{(i,t)}+1)
\\
&-
\psi(\alpha_{(i,t,v)}+1)
\bigr],
\end{split}
\label{eq:app-raw-entropy}
\end{equation}%
where \(\psi\) is the digamma function:
\begin{equation}
\psi(u)
=
\frac{d}{du}\log \Gamma(u)
=
\frac{\Gamma'(u)}{\Gamma(u)}.
\label{eq:app-digamma}
\end{equation}
The uncertainty \(H_k\) is small when the document-conditioned evidence is concentrated on a few next-token candidates, and larger when the evidence is spread across many candidates.

\paragraph{Token-Level Confidence Score}\leavevmode

The token-level confidence used in the main decoding rule is then
\begin{equation}
c_{(i,t)}
=
1-H_k(\mathbf{z}^i_t).
\label{eq:app-token-confidence}
\end{equation}
Here \(H_k\) is the expected entropy of \autoref{eq:app-raw-entropy}, with \(0\le H_k\le\log k\), and it is not rescaled by \(\log k\).
Thus, \(c_{(i,t)}\) is high when document \(d_i\) yields a sharp next-token prediction at step \(t\), and low when the model remains uncertain about how to continue from that document. Unlike the document-level confidence \(q_i\), which is computed once per question--document pair, \(c_{(i,t)}\) is recomputed at every decoding step.

\subsection{Logit Processing}
\label{app:algorithm}

In practice, \autoref{eq:dccd-rule} is applied to next-token log-probabilities, which differ from the logits by a per-stream constant and therefore leave the greedy choice unchanged. Each stream is first restricted to its top-$p$ nucleus ($p=0.95$), and the \texttt{<think>} and \texttt{</think>} tokens are masked because thinking is disabled. Tokens outside the nucleus of the positive or negative stream keep their full-context score, and tokens outside the full-context nucleus are excluded. The token-level confidence $c_{(i,t)}$ is computed from the raw logits, before this processing. The contrastive baselines use the same processing, which leaves full-context greedy decoding unchanged apart from the masked tags. Zero-shot answers use standard greedy decoding (\autoref{app:baselines}).

\subsection{Computational Cost}
\label{app:cost}

The document-level support probe is computed once per query. All $n$ document probes can be batched into a single forward pass, producing the $n$ values of $q_i$. This cost is amortized over the entire generation.

At each decoding step, \method requires one full-context forward stream and $n$ single-document forward streams. Thus, for $n$ retrieved documents, the decoding stage uses $n+1$ streams per token. The token-level confidence computation adds only $O(nk)$ operations for top-$k$ extraction and digamma evaluations, which is negligible compared with the model forward passes for the values of $k$ used in our experiments.

The implementation can use the model's standard KV cache for each stream. With caching, each decoding step only extends the existing streams by one token. The memory footprint and the computation per token scale linearly with the number of streams, on top of the one-time support-probe pass. The streams are decoded as one batch, so the measured runtime grows more slowly (Appendix~\ref{app:runtime}).

Two options reduce this cost. First, the Doc-Only variant (\S\ref{sec:ablations}) selects the positive and negative streams once from $q_i$, so it needs only three streams per step (the full context and two documents) regardless of $n$. It matches full \method on DRQA@5 but is weaker at @10. Second, reranking or filtering the bundle before decoding reduces $n$ for full \method. Hard filtering, however, cannot recover evidence that the filter discards.

\section{Experimental Details}
\label{app:experiments}

This appendix provides additional experimental details that describe the datasets, baselines, models, and headline evaluation protocol. Here, we record the implementation-level choices needed to reproduce the experiments, including dataset formatting, prompting, decoding configurations, and hyperparameters. 

\subsection{Dataset Details}
\label{app:datasets}

Table~\ref{tab:datasets-detail} summarizes the datasets used in our experiments. Natural Questions (NQ) and TriviaQA (TQA) use the preprocessing of \citet{izacard-grave-2021-leveraging}, which provides DPR-retrieved passages with title and text fields. PopQA uses the 4,267-question test split of a retrieval-augmented release of \citet{mallen-etal-2023-trust} (\texttt{Atipico1/popQA\_preprocessed} on Hugging Face), with ten retrieved passages per question. RetrievalQA follows \citet{zhang-etal-2024-retrievalqa} and is converted into the same document-list format. DRQA is our factual-conflict benchmark derived from DRBench, which consists of 568 internal enterprise facts.

\begin{table}[h]
\centering\small
\setlength{\tabcolsep}{6pt}
\begin{tabular}{@{}lccc@{}}
\toprule
\textbf{Dataset} & \textbf{\# Examples} & \textbf{Avg @5} & \textbf{Avg @10} \\
\midrule
NQ           & 3{,}610  &   526 & 1{,}042 \\
TQA          & 11{,}313 &   527 & 1{,}041 \\
PopQA        & 4{,}267  &   521 & 1{,}036 \\
RetrievalQA  & 2{,}785  &   443 & 1{,}003 \\
DRQA         & 568      & 2{,}039 & 4{,}002 \\
\bottomrule
\end{tabular}
\caption{Dataset statistics used in our experiments. We include the average number of words of the input for the @5 and @10 settings.}
\label{tab:datasets-detail}
\end{table}

\subsection{Prompt Templates}
\label{app:prompting}

All methods use the same answer-generation prompt format. We adopt the same prompt template from prior work \citep{shi-etal-2024-trusting, jin-etal-2024-dvd} with the zero-shot and contextualized prompts displayed below.

\noindent Zero-shot Prompt:
\begin{Prompt}
Question: {question}
Answer:
\end{Prompt}

\noindent Contextualized Prompt:
\begin{Prompt}
Write a high-quality concise answer for the given question using only the provided search results (some of which might be irrelevant). 

{search_results}

Question: {question}
Answer:
\end{Prompt}

Both prompts end with a single space after ``Answer:'', as does the support probe (\autoref{app:support-probe}). The \texttt{\{search\_results\}} slot lists the documents one per line as \texttt{Document [$i$](Title: $t_i$) $d_i$}, where DRQA documents use the placeholder title ``Doc~$i$''. Each per-document prompt and each support probe contains one document rendered the same way, with its original index~$i$.

All prompts are wrapped with the default chat template provided by the corresponding Hugging Face tokenizer, with thinking disabled for \texttt{Qwen3.5}. We use the chat template both for answer generation and for the document-level support probe in \method, so confidence estimation and final decoding are performed under the same instruction-following format. The per-document prompts follow the same template as the contextualized prompt following \citet{jin-etal-2024-dvd}.

\subsection{Baseline Decoding Configurations}
\label{app:baselines}

We use greedy decoding with temperature $0$ and a maximum generation length of $60$ new tokens for all methods. To reduce artifacts from recall-oriented string-match metrics, we generate and evaluate a single-sentence answer. Since \texttt{str-em} rewards the presence of a normalized gold string anywhere in the prediction, long generations that copy retrieved passages or list multiple candidate answers can be artificially rewarded even when the model does not clearly commit to the correct answer. Single-sentence generation makes the comparison across decoding methods more meaningful by forcing each method to produce a concise prediction. The first sentence is defined in Appendix~\ref{app:eval}. Stopping generation once it is complete gives the same sentence as generating all 60 tokens. Unless otherwise stated, hyperparameters are fixed across all datasets and models. We describe the hyperparameter settings for all decoding strategies used in our experiments and summarize in \autoref{tab:hparams-summary}.

\paragraph{Zero-shot Decoding}
The zero-shot baseline generates from the question alone, without any retrieved documents. It is included to estimate how much of each dataset can be answered from parametric memory alone. It uses standard greedy decoding (Hugging Face \texttt{generate}) on the zero-shot prompt, without the stream processing of Appendix~\ref{app:algorithm}.

\paragraph{Full-Context Decoding}
The full-context baseline decodes directly from the logits conditioned on the question and the full retrieved document set:
\[
\mathbf{z}_t = \mathbf{z}^{\mathrm{full}}_t.
\]

\paragraph{CAD}
Context-Aware Decoding (CAD) contrasts full-context logits against no-context logits:
\[
\mathbf{z}^{\mathrm{CAD}}_t
=
(1+\alpha)\mathbf{z}^{\mathrm{full}}_t
-
\alpha \mathbf{z}^{\mathrm{none}}_t.
\]
We use $\alpha=0.2$ based on the results reported by \citet{shi-etal-2024-trusting}.

\paragraph{\textsc{AdaCAD}}
Adaptive Context Aware Decoding (\textsc{AdaCAD}) replaces the fixed CAD weight with a token-level adaptive weight based on the Jensen-Shannon divergence between the full-context and no-context distributions:
\[
\alpha_t
=
\mathrm{JSD}
\left(
p^{\mathrm{full}}_t
\Vert
p^{\mathrm{none}}_t
\right),
\]
\[
\mathbf{z}^{\mathrm{\textsc{AdaCAD}}}_t
=
(1+\alpha_t)\mathbf{z}^{\mathrm{full}}_t
-
\alpha_t\mathbf{z}^{\mathrm{none}}_t.
\]
We set the minimum weight floor to $0.0$ following \citet{wang-etal-2025-adacad}.

\paragraph{DVD}
Dynamic Contrastive Decoding (DVD) is a document-aware contrastive decoding baseline for multi-document QA. We use the static-weight configuration with $\beta=0.25$, $\gamma=0.2$, and document top-$k=10$. In early experiments, we find that dynamic $\beta$ performs poorly, so we followed the suggested configuration from \citet{jin-etal-2024-dvd}. The method selects document-conditioned streams according to its entropy-based criterion and applies a document-level contrast together with a context-prior contrast.

\paragraph{\textsc{CoCoA}}
Confidence- and Context-Aware Adaptive Decoding (\textsc{CoCoA}) uses a confidence-aware interpolation between no-context and full-context predictions. We use the paper-faithful configuration with Rényi divergence order $\alpha=0.5$, entropy-gap weight $\gamma=1.0$, peakedness amplifier $z=5.0$, and numerical stability constants $\delta=10^{-8}$ and $\varepsilon=10^{-12}$ following \citet{khandelwal-etal-2025-cocoa}.

\paragraph{\method}
Our proposed \methodlong (\method) has the advantage of being relatively hyperparameter-free, with the only hyperparameter being the top-$k$ logits used for computing token-level confidence, which we set to $k=10$ throughout all our experiments.

\begin{table}[h]
\centering
\small
\setlength{\tabcolsep}{4pt}
\begin{tabular}{@{}lll@{}}
\toprule
\textbf{Method} & \textbf{Hyperparameter} & \textbf{Value} \\
\midrule
All methods & Temperature & $0$ \\
All methods & Decoding strategy & greedy \\
All methods & Max new tokens & $60$ \\
All methods & Top-$p$ nucleus$^\dagger$ & $0.95$ \\
CAD & $\alpha$ & $0.2$ \\
\textsc{AdaCAD} & JSD floor & $0.0$ \\
DVD & $\beta$ & $0.25$ \\
DVD & $\gamma$ & $0.2$ \\
DVD & Document top-$k$ & $10$ \\
\textsc{CoCoA} & Rényi order & $0.5$ \\
\textsc{CoCoA} & Entropy-gap weight & $1.0$ \\
\textsc{CoCoA} & Peakedness amplifier & $5.0$ \\
\textsc{CoCoA} & Numerical floors & $\delta=10^{-8}$, $\varepsilon=10^{-12}$ \\
\method & top-$k$ & 10 \\
\bottomrule
\end{tabular}
\caption{Main decoding hyperparameters used in the experiments. Hyperparameters are fixed across datasets and models. $^\dagger$Per-stream nucleus applied before combining (Appendix~\ref{app:algorithm}), not used by zero-shot decoding.}
\label{tab:hparams-summary}
\end{table}

\subsection{Models and Inference Backend}
\label{app:models}

We evaluate the three instruction-tuned language models  through Hugging Face \texttt{transformers} \citep{wolf-etal-2020-transformers}, namely, \texttt{Qwen3.5-2B}\footnote{\noindent \url{https://huggingface.co/Qwen/Qwen3.5-2B}}, \texttt{Qwen3.5-9B}\footnote{\noindent \url{https://huggingface.co/Qwen/Qwen3.5-9B}}, and \texttt{Phi-3-medium}\footnote{\noindent \url{https://huggingface.co/microsoft/Phi-3-medium-128k-instruct}}. All models are loaded with their default tokenizers and chat templates. We use bfloat16 inference when supported. Tokenizer padding is set to the left so the final-position logits are aligned across batched streams.

For contrastive decoding methods, each decoding step may require multiple forward streams. CAD, \textsc{AdaCAD}, and \textsc{CoCoA} use no-context and full-context streams. DVD and \method additionally use document-conditioned streams. For \method, the support probe is computed once before decoding, while the full-context and document-conditioned logits are computed during each decoding step. Streams are batched whenever possible. In our implementation, every method except zero-shot advances the same left-padded batch of $n+2$ streams (no-context, full-context, and the $n$ single-document prompts) with a KV cache and combines only the streams it needs, so all methods see numerically identical streams.

\subsection{Evaluation Metrics}
\label{app:eval}

\paragraph{Normalized string exact match.}
For NQ, TriviaQA, PopQA, and RetrievalQA, we use normalized string exact match, denoted \texttt{str-em}. We first extract the generated answer sentence, i.e., the text up to the first line break or up to the first period that follows three lowercase letters and precedes a space and an uppercase letter (leading lines without letters are skipped), and then normalize both the prediction and gold answers using the standard QA normalization procedure: lowercasing, removing punctuation, removing articles, and collapsing whitespace. A prediction is counted as correct if any normalized gold answer matches the normalized prediction by exact or substring match, following common retrieval-augmented QA evaluation practice.

\paragraph{LLM-as-a-judge for DRQA.}
For DRQA, the reference answer is a free-form enterprise insight rather than a short canonical span. We therefore evaluate semantic correctness with an LLM judge. The judge receives the question, the reference answer, and the first sentence of the model prediction, and returns a binary correctness label. A prediction is marked correct if it conveys the same answer as the reference at the required level of specificity. Numerical and date answers must match the reference detail; refusals are marked incorrect unless the reference answer itself indicates that the information is unavailable.

We use deterministic judge decoding with temperature $0$. Judge outputs are parsed as JSON with a boolean correctness field. If the judge response cannot be parsed, the example is conservatively marked incorrect. Below is the prompt template for our judge using \texttt{anthropic/claude-opus-4-5} \cite{anthropic2025claudeopus45cr} through OpenRouter API\footnote{\url{https://openrouter.ai}}.

\begin{Prompt}
You are a strict but fair grader for short-answer
question answering. Decide whether the model answer
matches the reference answer.

Question:
{question}

Reference answer:
{reference_answer}

Model answer:
{prediction}

Mark the model answer correct if and only if it
conveys the meaning of the reference answer.

Rules:
1. Surface form does not matter if the meaning is
   equivalent.
2. Numerical and date answers must match the
   reference at the level of detail provided.
3. A refusal or "not specified" answer is incorrect
   unless the reference answer also says the
   information is unavailable.
4. Extra context is acceptable only if the answer
   clearly commits to the correct value.
5. If the answer gives a wrong primary value but
   mentions the correct value only incidentally,
   mark it incorrect.

Return exactly one JSON object:
{"correct": true_or_false, "reason": "<short reason>"}
\end{Prompt}

\subsection{Hardware and Runtime}
\label{app:runtime}

All experiments can be run on a single NVIDIA H100 80GB GPU. Runtime depends on model size, dataset size, retrieval depth, and decoding method. Full-context decoding advances a single stream. CAD, \textsc{AdaCAD}, and \textsc{CoCoA} add the no-context stream. DVD and \method also advance one stream per retrieved document, and \method runs the document-level support probe once per question. \autoref{tab:runtime} reports the measured wall-clock runtime.

\begin{table}[ht]
\centering
\resizebox{\linewidth}{!}{%
\setlength{\tabcolsep}{5pt}\begin{tabular}{@{}lrr|rr|rr|rr@{}}
\toprule
& \multicolumn{2}{c|}{\textbf{Full}} & \multicolumn{2}{c|}{\textbf{CAD family}} & \multicolumn{2}{c|}{\textbf{DVD}} & \multicolumn{2}{c}{\textbf{\method}} \\
\textbf{Dataset} & @5 & @10 & @5 & @10 & @5 & @10 & @5 & @10 \\
\midrule
DRQA         & 11  & 14  & 13  & 16  & 16  & 31  & 19  & 35  \\
RetrievalQA  & 36  & 34  & 39  & 36  & 42  & 52  & 43  & 63  \\
NQ           & 65  & 65  & 67  & 69  & 72  & 87  & 82  & 101 \\
PopQA        & 53  & 57  & 53  & 62  & 64  & 88  & 73  & 98  \\
TriviaQA     & 155 & 171 & 158 & 200 & 191 & 226 & 214 & 284 \\
\bottomrule
\end{tabular}
}
\caption{Wall-clock runtime in minutes for one pass over each dataset with \texttt{Qwen3.5-2B} on a single NVIDIA H100 80GB GPU. We time the same 20 randomly sampled questions for every method and retrieval depth and scale the mean time per question to the size of the dataset, which leaves a sampling error of about 10\%. Generation stops after the first sentence. Full-context decoding advances one stream. The CAD family (CAD, \textsc{AdaCAD}, and \textsc{CoCoA}, averaged) advances two, and DVD and \method advance all $n+2$. This is the \texttt{minimal\_streams} option of the released code. By default the code advances all $n+2$ streams for every method except zero-shot (Appendix~\ref{app:models}), which makes full-context decoding about as slow as DVD.}
\label{tab:runtime}
\end{table}

For $n$~retrieved documents, \method advances one stream for the full retrieved context and $n$ streams for the individual documents, so the computation per generated token grows linearly with $n$. Wall-clock time grows more slowly because the streams are decoded as one batch. In \autoref{tab:runtime}, \method takes about 1.2 to 1.7 times as long as full-context decoding at @5 and about 1.5 to 2.5 times as long at @10. The token-level confidence computation is inexpensive relative to the model forward pass because it only requires top-$k$ extraction and digamma evaluations over a small number of tokens. The document-level support probe is computed once per question--document pair and is amortized over the generated answer.

\section{Qualitative Examples}
\label{app:qualitative}

We present representative qualitative examples for \texttt{Qwen3.5-2B} with
top-10 retrieved documents. Each example shows the answer sentence scored for each decoding method (\autoref{app:eval}). It does not end at a period preceded by fewer than three lowercase letters, as in ``Andaman Sea.''. Predictions are truncated to 220 characters when
necessary. Correct predictions are shaded green and incorrect predictions are
shaded red.

For NQ, TriviaQA, RetrievalQA, and PopQA, correctness follows the first-sentence
\texttt{str-em} metric described in \autoref{app:eval}. For DRQA, correctness is
determined by the LLM-as-a-judge metric. To avoid \texttt{str-em} false positives
from long generations that enumerate multiple candidate answers, we additionally
filter qualitative examples to predictions of at most 240 characters that commit
to a single value when the gold answer is numeric and contain no refusal phrase
such as \emph{``no information''} or \emph{``cannot determine''}.

\newcommand{\correctrow}[2]{%
\cellcolor{posgreen}\textbf{#1} & \cellcolor{posgreen}#2 \\
}
\newcommand{\wrongrow}[2]{%
\cellcolor{negred}#1 & \cellcolor{negred}#2 \\
}

\newcommand{\exampleheader}[3]{%
\medskip
\noindent\textbf{Q (#1):} \emph{#2} \\
\textbf{Gold:} #3
\medskip
}

\newenvironment{predictiontable}{%
\begin{center}
\small
\setlength{\tabcolsep}{5pt}
\renewcommand{\arraystretch}{1.08}
\begin{tabular}{@{}p{0.17\linewidth}p{0.78\linewidth}@{}}
\toprule
\textbf{Method} & \textbf{Prediction} \\
\midrule
}{%
\bottomrule
\end{tabular}
\end{center}
}

\subsection{DRQA}

\exampleheader{qid 194}
{How did Jason Wong respond to a simulated phishing incident in March 2025?}
{In March 2025, Jason Wong coordinated with the IT helpdesk and quickly contained a simulated phishing attack, reducing average internal response time to under 25 minutes.}

\begin{predictiontable}
\wrongrow{full}{``Based on the provided documents, there is a discrepancy regarding the specific date of the incident.''}
\wrongrow{CAD}{``Based on the provided documents, there is a discrepancy regarding the specific month of the incident:''}
\wrongrow{\textsc{AdaCAD}}{``Based on the provided documents, there is a discrepancy regarding the specific month of the incident:''}
\wrongrow{DVD}{``Based on the provided documents, there is a discrepancy regarding the specific month of the incident and the outcome:''}
\wrongrow{\textsc{CoCoA}}{``Based on the provided search results, there is a contradiction regarding the specific details of Jason Wong's response to a simulated phishing incident in March 2025.''}
\correctrow{\method}{``Jason Wong responded to the simulated phishing incident in March 2025 by coordinating with the IT helpdesk to contain the threat expeditiously.''}
\end{predictiontable}

\subsection{Natural Questions}

\exampleheader{qid 29}
{in which sea pearl is found in india}
{the Indian Ocean}

\begin{predictiontable}
\wrongrow{full}{``Based on the provided search results, the sea where pearls are found in India is the Andaman Sea.''}
\wrongrow{CAD}{``Based on the provided search results, India's major mineral resources include tin deposits off the coasts of Malaysia and Thailand, which are located in the Andaman Sea. Additionally, the Andaman Sea is noted...''}
\wrongrow{\textsc{AdaCAD}}{``Based on the provided search results, India's major mineral resources include tin deposits off the coasts of Malaysia and Thailand, which are located in the Andaman Sea. Additionally, the Andaman Sea is noted...''}
\wrongrow{DVD}{``Based on the provided search results, the sea pearl found in India is the String of Pearls.''}
\wrongrow{\textsc{CoCoA}}{``The sea!''}
\correctrow{\method}{``Based on the provided search results, the sea where pearls are found in India is the Indian Ocean.''}
\end{predictiontable}

\exampleheader{qid 1491}
{who wrote it's gonna take a lot of love}
{Neil Young}

\begin{predictiontable}
\wrongrow{full}{``Based on the provided search results, there is no mention of a song titled ``It's gonna take a lot of love'' or its author.''}
\wrongrow{CAD}{``Based on the provided search results, there is no document that states who wrote the song ``It's Gonna Take a Lot of Love.''''}
\wrongrow{\textsc{AdaCAD}}{``Based on the provided search results, there is no document that explicitly states who wrote the song ``It's Gonna Take a Lot of Love.''''}
\wrongrow{DVD}{``Based on the provided search results, there is no document that states who wrote the song ``It's Gonna Take a Lot of Love.''''}
\wrongrow{\textsc{CoCoA}}{``The song ``It's Gonna Take a Lot of Love'' was written by Linda Ronstadt and Lotta Love.''}
\correctrow{\method}{``The song ``Lotta Love'' was written and recorded by Neil Young.''}
\end{predictiontable}

\exampleheader{qid 2783}
{who is the all time leading scorer in ncaa tournament history}
{Pete Maravich}

\begin{predictiontable}
\wrongrow{full}{``Based on the provided search results, the all-time leading scorer in NCAA tournament history is Gary Gait, who holds the record for the most goals in the NCAA tournament with 16 goals.''}
\wrongrow{CAD}{``Based on the provided search results, Phil Latreille is the all-time leading scorer in NCAA Division I history with 4,045 points.''}
\wrongrow{\textsc{AdaCAD}}{``Based on the provided search results, the all-time leading scorer in NCAA tournament history is Gary Gait, who holds the record for the most goals in the NCAA tournament with 16 goals.''}
\wrongrow{DVD}{``Based on the provided search results, Don MacLean is the all-time leading scorer in NCAA tournament history, having scored 2,608 points.''}
\wrongrow{\textsc{CoCoA}}{``The all-time leading scorer in NCAA tournament history is Matt Ward.''}
\correctrow{\method}{``Based on the provided search results, the answer is Pete Maravich.''}
\end{predictiontable}

\exampleheader{qid 2792}
{where do they film young and the restless}
{CBS Television City}

\begin{predictiontable}
\wrongrow{full}{``The series ``The Young and the Restless'' was originally filmed in Los Angeles, California, with production taking place there while the show aired in New York.''}
\wrongrow{CAD}{``The show ``The Young and the Restless'' was originally filmed in Los Angeles, California, with production taking place there while writers worked from Chicago.''}
\wrongrow{\textsc{AdaCAD}}{``The show ``The Young and the Restless'' was originally filmed in Los Angeles, California, with production taking place there while writers worked from Chicago.''}
\wrongrow{DVD}{``Based on the provided search results, the soap opera ``The Young and the Restless'' was filmed in Los Angeles, California.''}
\wrongrow{\textsc{CoCoA}}{``The 2017 film \emph{Young and the Restless} was filmed in Los Angeles, California.''}
\correctrow{\method}{``They film ``The Young and the Restless'' at CBS Television City, specifically studios 41 and 43 in Hollywood.''}
\end{predictiontable}

\exampleheader{qid 1600}
{where does stevia in the raw come from}
{the plant species \emph{Stevia rebaudiana}}

\begin{predictiontable}
\wrongrow{full}{``Stevia In The Raw is a zero-calorie sweetener developed by Cumberland Packing Corporation.''}
\wrongrow{CAD}{``Stevia in the Raw is a stevia-based low-calorie sweetener developed by Cumberland Packing Corporation.''}
\wrongrow{\textsc{AdaCAD}}{``Stevia in the Raw is a stevia-based low-calorie sweetener developed by PureVia, a product created jointly by PepsiCo and Whole Earth Sweetener Company.''}
\wrongrow{DVD}{``Stevia in the Raw is a sweetener developed by Cumberland Packing Corporation that is made from stevia leaf extract, erythritol, and natural flavors.''}
\wrongrow{\textsc{CoCoA}}{``Stevia In The Raw is a dietary supplement that!''}
\correctrow{\method}{``Stevia in the Raw comes from the leaves of the plant species \emph{Stevia rebaudiana}, native to Brazil and Paraguay.''}
\end{predictiontable}

\subsection{TriviaQA}

\exampleheader{qid 7277}
{The name of a knot on a lasso is also the name of a Japanese motor manufacturer. What is the name?}
{HONDA}

\begin{predictiontable}
\wrongrow{full}{``The name is Miyata.''}
\wrongrow{CAD}{``The name is Weds, or WEDS.''}
\wrongrow{\textsc{AdaCAD}}{``Miyata''}
\wrongrow{DVD}{``The name is Weds, or WEDS.''}
\wrongrow{\textsc{CoCoA}}{``The name is !WORK.''}
\correctrow{\method}{``The name is Honda.''}
\end{predictiontable}

\exampleheader{qid 8593}
{In imperial measure, how many gills are there in a gallon?}
{32}

\begin{predictiontable}
\wrongrow{full}{``There are 16 imperial gills in an imperial gallon.''}
\wrongrow{CAD}{``There are 16 imperial gills in an imperial gallon.''}
\wrongrow{\textsc{AdaCAD}}{``There are 16 imperial gills in an imperial gallon.''}
\wrongrow{DVD}{``There are 160 imperial gills in an imperial gallon.''}
\wrongrow{\textsc{CoCoA}}{``In the imperial system, there are 16! (sixteen) gills in one gallon.''}
\correctrow{\method}{``In the imperial system, a gallon is equal to 32 gills.''}
\end{predictiontable}

\exampleheader{qid 4705}
{How many US states border Lake Superior?}
{three}

\begin{predictiontable}
\wrongrow{full}{``Based on the provided documents, two U.S. states border Lake Superior: Minnesota and Wisconsin.''}
\wrongrow{CAD}{``Based on the provided documents, two U.S. states border Lake Superior: Michigan and Wisconsin.''}
\wrongrow{\textsc{AdaCAD}}{``Based on the provided documents, two U.S. states border Lake Superior: Minnesota and Wisconsin.''}
\wrongrow{DVD}{``Based on the provided search results, two U.S. states border Lake Superior: Minnesota and Wisconsin.''}
\wrongrow{\textsc{CoCoA}}{``There are 4 US states that border Lake Superior:''}
\correctrow{\method}{``Three U.S. states border Lake Superior: Minnesota, Wisconsin, and Michigan.''}
\end{predictiontable}

\exampleheader{qid 5136}
{In which ``Star Wars'' episode does Anakin Skywalker transform into the black-suited Darth Vader?}
{\emph{Star Wars: Episode III -- Revenge of the Sith}}

\begin{predictiontable}
\wrongrow{full}{``Anakin Skywalker transforms into the black-suited Darth Vader in the ``Star Wars'' episode ``Star Wars: Episode IV -- A New Hope,'' which was released in 1977.''}
\wrongrow{CAD}{``Based on the provided search results, Anakin Skywalker transforms into the black-suited Darth Vader in the animated series ``Star Wars: Clone Wars.''''}
\wrongrow{\textsc{AdaCAD}}{``Based on the provided search results, Anakin Skywalker transforms into the black-suited Darth Vader in the animated series ``Star Wars: Clone Wars.''''}
\wrongrow{DVD}{``Based on the provided search results, the specific episode where Anakin Skywalker transforms into Darth Vader is not explicitly named as a single title in the text.''}
\wrongrow{\textsc{CoCoA}}{``![](https://![](https://![](https://![]...''}
\correctrow{\method}{``Anakin Skywalker transforms into the black-suited Darth Vader in ``Star Wars: Episode III -- Revenge of the Sith.''''}
\end{predictiontable}

\exampleheader{qid 8453}
{What most-rejected 1974 bestseller describes the 17-day journey of the author, who calls himself Phaedrus, and his son Chris from Minnesota to California?}
{\emph{Zen and the Art of Motorcycle Maintenance} by Robert M. Pirsig}

\begin{predictiontable}
\wrongrow{full}{``The bestseller is \emph{Zen and the Art of Motorcycle Maintenance} by Robert Pirsig.''}
\wrongrow{CAD}{``The 1974 bestseller that describes the 17-day motorcycle journey from Minnesota to Northern California, authored by Richard Pirsig and fictionalized as an autobiography of Phaedrus and his son Chris, is...''}
\wrongrow{\textsc{AdaCAD}}{``Based on the provided search results, there is no bestseller that describes an 17-day journey from Minnesota to California by an author named Phaedrus.''}
\wrongrow{DVD}{``The bestseller is \emph{Zen and the Art of Motorcycle Maintenance} by Robert Pirsig.''}
\wrongrow{\textsc{CoCoA}}{``The book you are referring to is \emph{The Road to California}!''}
\correctrow{\method}{``The bestseller is \emph{Zen and the Art of Motorcycle Maintenance} by Robert M. Pirsig.''}
\end{predictiontable}

\end{document}